\documentclass{article}

     \PassOptionsToPackage{numbers, compress}{natbib}

 \usepackage{iclr2027_conference,times}

\usepackage[utf8]{inputenc} 
\usepackage[T1]{fontenc}    
\usepackage{hyperref,algorithm,algorithmic,bm}       
\usepackage{url}            
\usepackage{booktabs}       
\usepackage{amsfonts}       
\usepackage{nicefrac}       
\usepackage{microtype}      
\usepackage[dvipsnames]{xcolor}  

\hypersetup{
    colorlinks=true, 
    linkcolor=Blue,
    filecolor=Blue,      
    urlcolor=Blue,
    citecolor=Blue 
    }

\usepackage{enumitem}
\setlist[itemize]{noitemsep, topsep=0pt}
\setlist[enumerate]{noitemsep, topsep=0pt}

\usepackage{adjustbox}
\usepackage{tabularray}
\UseTblrLibrary{booktabs}
\usepackage{array}
\newcolumntype{/}{!{\vrule width 1pt}}

\usepackage{amsmath}
\usepackage{amssymb}
\usepackage{mathtools}
\usepackage{amsthm}
\usepackage{dsfont} 
\usepackage{bbm}    

\newcommand{\ind}{\perp\!\!\!\perp}

\theoremstyle{plain}
\newtheorem{theorem}{Theorem}[section]
\newtheorem{proposition}[theorem]{Proposition}
\newtheorem{lemma}[theorem]{Lemma}

\theoremstyle{definition}
\newtheorem{definition}[theorem]{Definition}
\newtheorem{assumption}[theorem]{Assumption}
\theoremstyle{remark}
\newtheorem{remark}[theorem]{Remark}
\theoremstyle{example}

\usepackage{tikz}
\usepackage{pgfplots,pgfmath,pgffor}
\usepackage{tikz-3dplot} 
\usetikzlibrary {arrows.meta,arrows,patterns,decorations.pathreplacing,chains,positioning,shapes.geometric,shapes.callouts,calc,backgrounds,trees,pgfplots.fillbetween,quotes,calligraphy,automata,fit}

\usepackage{ragged2e}

\usepackage{wrapfig}

\usepackage[most]{tcolorbox}
\AtBeginEnvironment{tcolorbox}{\footnotesize}
    { 
    \end{tcolorbox}
    }
    { 
    \end{tcolorbox}
    }

\newcommand{\convexpath}[2]{
[   
    create hullnodes/.code={
        \global\edef\namelist{#1}
        \foreach [count=\counter] \nodename in \namelist {
            \global\edef\numberofnodes{\counter}
            \node at (\nodename) [draw=none,name=hullnode\counter] {};
        }
        \node at (hullnode\numberofnodes) [name=hullnode0,draw=none] {};
        \pgfmathtruncatemacro\lastnumber{\numberofnodes+1}
        \node at (hullnode1) [name=hullnode\lastnumber,draw=none] {};
    },
    create hullnodes
]
($(hullnode1)!#2!-90:(hullnode0)$)
\foreach [
    evaluate=\currentnode as \previousnode using \currentnode-1,
    evaluate=\currentnode as \nextnode using \currentnode+1
    ] \currentnode in {1,...,\numberofnodes} {
  let
    \p1 = ($(hullnode\currentnode)!#2!-90:(hullnode\previousnode)$),
    \p2 = ($(hullnode\currentnode)!#2!90:(hullnode\nextnode)$),
    \p3 = ($(\p1) - (hullnode\currentnode)$),
    \n1 = {atan2(\y3,\x3)},
    \p4 = ($(\p2) - (hullnode\currentnode)$),
    \n2 = {atan2(\y4,\x4)},
    \n{delta} = {-Mod(\n1-\n2,360)}
  in 
    {-- (\p1) arc[start angle=\n1, delta angle=\n{delta}, radius=#2] -- (\p2)}
}
-- cycle
}

\usepackage{pifont}

\definecolor{platinum}{rgb}{0.9, 0.89, 0.89}

\usepackage[toc,page,header]{appendix}
\usepackage{minitoc}

\title{From Causal Graphs to Markovian States: Multi-Order State Exposure for Deep RL}

\author{
    Jiamin Xu \thanks{Corresponding author. Email: jx389@cornell.edu}  \\
    Cornell Tech, New York, NY
    \And
    Jacqueline Maasch \\
    Cornell Tech, New York, NY
    \And
    Kyra Gan  \thanks{Corresponding author. Email: jg2424@cornell.edu}\\
    Cornell Tech, New York, NY
} 
\iclrfinalcopy

\begin{document}

\doparttoc 
\faketableofcontents 

\maketitle
\lhead{}

\begin{abstract}
Online reinforcement learning (RL) relies on the Markov property for guaranteed performance, but real-world applications often lack well-defined states given raw observed variables. While causal RL has attracted growing interest, existing work typically assumes Markovian states are provided and uses causality only to accelerate learning.
This leaves a fundamental gap: \emph{given a longitudinal causal graph over observed variables, how does one construct MDP states that provably satisfy the Markov property?} We provide
a procedure that constructs a provably minimal such state. 
In deep RL, 
the minimal representation alone
empirically fails to improve performance,
indicating that neural networks cannot automatically exploit Markovian minimality. 
We therefore propose \textbf{MOSE} (Multi-Order State Exposure), which
exposes a single shared Q-function to history-based state constructions of orders 1 through $W$ simultaneously.
An attention-based selection layer scores the dimensions of this combined input
by their estimated relevance to value-function estimation and selects the high-scoring ones.
Causal MOSE further 
constrains the selection layer to retain every variable of the minimal state, guaranteeing sufficiency while leaving the model free to admit useful redundancy.
MOSE consistently outperforms both the minimal state construction and single-window policies on common benchmarks and synthetic datasets. Including the minimal representation alongside MOSE can further improve performance. Our results establish a core principle for causal deep RL: minimal sufficiency is not enough. \emph{Controlled redundancy} gives the model flexibility to learn its state representation end-to-end. Causal state information provides structural guidance that can make this learning more efficient.
\end{abstract}

\section{Introduction}
Reinforcement learning (RL) algorithms typically assume that the agent's state satisfies the Markov property: the future is conditionally independent of the past given the present \citep{sutton1998rl}. In practice, however, raw observations (e.g., sensor streams) do not inherently form a Markovian state, because the system’s dynamics may depend on an unknown amount of history. 
For instance, sepsis treatment depends on vital-sign trajectories, not just current readings \citep{komorowski2018artificial}; similarly, robotic manipulation may require tracking occluded objects across frames \citep{chung2026rethinking}. Using non-Markovian observations as states can, in theory, lead to suboptimal policies and unsafe decisions.

A large body of prior work addresses this challenge under the partially observable Markov decision process (POMDP) framework, where a latent Markovian state exists but is not directly observed \citep{monahan1982state}. Methods in this line of work learn a belief state using sequential architectures such as LSTMs \citep{hausknecht2015deep,kapturowski2018recurrent,badia2020agent57} or Transformers \citep{zambaldi2019deep,chen2021decision,zheng2022online}.
Though in principle these approaches can recover the optimal policy given sufficient history, they offer no guidance on which past observations are actually needed. Instead, this is learned from data, often requiring large samples. Further, it is unclear if the learned representation satisfies Markovian Property which makes state selection on the learned representation necessary. Meanwhile, \emph{frame stacking} arises as a pragmatic heuristic that has become widespread in deep RL practice.  Concretely, frame stacking concatenates a fixed number $k$ of recent observations into a single input, reducing partial observability and rendering the process approximately Markovian over $k$ time steps \citep{shi2020does}. Frame stacking can be effective when dependencies span a short, known horizon. For example, stacking four frames in Atari is assumed to sufficiently capture object motion and works well in practice \citep{mnih2015human,hessel2018rainbow,kaiser2020model}. 
However, as the true window length $k$ is often unknown, $k$ must be tuned or learned. When $k$ is too small, we lose critical information, while arbitrarily large $k$ requires excess computation and memory and leads to poor sample efficiency \citep{tasse2025finding}.

To move beyond these heuristics when constructing the state space, a principled approach would need to determine exactly which past variables are causally relevant for the future. 
While causal \emph{directed acyclic graphs} (DAGs) have been used in RL to improve learning, these DAGs are typically static (non‑longitudinal) and assume a pre‑specified, Markovian state representation. 
These methods do not ask the more basic question: \emph{how does one construct Markovian states from raw observations?} 
In this work, we bridge this gap by establishing a direct mapping from a known longitudinal causal graph to a \textit{minimal Markovian state representation}: a representation that satisfies the Markov property and is minimal in the sense that removing any variable would, for some consistent environment, either break the Markov property or reduce the optimal value. This establishes the long‑missing connection from longitudinal causal graphs to Markov decision process (MDP) states. 

While prior causal RL methods assume access to a valid MDP \citep{li2025automatic,li2025confounding}, 
we instead ask whether the state we construct can itself improve learning. The answer, however, is not straightforward in deep RL. Deep networks are known to overfit to spurious correlations and rely on redundant features which affect the expressivity of neural networks \citep{zhang2018study,nikishin2022primacy,sokar2023dormant,cobbe2019quantifying}. 
By stripping away all statistical slack, a minimal representation can lead to sharp minima and poor generalization \citep{zhang2018study,nikishin2022primacy,sokar2023dormant,cobbe2019quantifying}. Indeed, in our experiments, we discover that feeding only the minimal representation often fails to improve performance and can even degrade it. 

Motivated by this, we reconsider the role of redundancy. We propose MOSE (Multi‑Order State Exposure): we add a selective layer that will operate on the belief history either learned by sequential architecture or the raw observation that is markovian ($W^{th}$ order) so the model can select the redundant information which helps learning value function. To calculate importance score, we calculate the attention between a learnable query and each element of the history and then compare with a learnable threshold to decide whether to include that element as input or not.
This design is supported by a range of findings: (i) attention can calculate the relationship between elements \citep{zambaldi2019deep,vaswani2017attention}; (ii) exploration signals improve deep RL performance \citep{fortunato2018noisy,laskin2020reinforcement}; and (iii) multi-view learning helps neural networks build richer representations \citep{li2019multi,chen2021unsupervised,jangir2022look,hwang2023information}. Empirically, MOSE substantially outperforms both the minimal state and standard frame stacking in multiple environments. Moreover, we introduce Causal-MOSE, which always inputs minimal markovian state while the model can decide whether to retain other information. Causal-MOSE performed as well or better than MOSE in experiments, suggesting that controlled redundancy can unlock the benefit of causal state information.

\textbf{Contributions \;} This work presents three main contributions.
\begin{enumerate}[leftmargin=*]
    \item \textit{Minimal Markovian state construction from time series causal DAGs.} We prove a graphical criterion for constructing a valid state space given a time series causal DAG. We provide a procedure that returns a provably minimal  state space satisfying validity (Algorithm \ref{alg:backward_state_construction}).
    \item \textit{Multi-Order State Exposure (MOSE).} We provide a heuristic method for state space construction that does not require knowledge of the causal DAG (Algorithm \ref{alg:mose}). We show that in deep RL, MOSE improves upon both the naive frame stacking heuristic and the minimal markovian state provided by Algorithm \ref{alg:backward_state_construction}.
    \item \textit{Causal-MOSE.} Finally, we show that combining the minimal Markovian state with MOSE performs as well or better on the experimental settings explored in this work.
\end{enumerate}

\section{Related Works}

This work lies at the intersection of RL, causality, and deep learning. See Appendix~\ref{appendix:related_works} for extended related works.

\textbf{Markovian Tests}\;\; Non-Markovianity is a known issue in RL, yet testing it requires new test statistics as classical conditional independence tests (CITs) assume i.i.d. data. Recent efforts have addressed this by developing asymptotically valid test procedures (e.g., conditional characteristic‑function tests \citep{chen2012testing} and forward‑backward sequential tests \citep{shi2020does}), embedding such tests in deep generative frameworks for high‑dimensional settings \citep{zhou2023testing}. 
These Markovian tests generalize CITs to time series, enabling us to directly test whether a future variable is conditionally independent of the full past given a candidate parent set. We further discuss tests of Markovianity in our framework in Section~\ref{subsec:prelim}. 

\textbf{Causal Reinforcement Learning}\;\;
A growing literature explores the intersection of causality and RL. Assuming access to a ground truth DAG, causal graphs have been explored in \emph{reward design} for mediator-based surrogate rewards~\citep{zou2025causal}, robust reward modeling~\citep{srivastava2025robust}, and causal Bellman equations under confounded offline data \citep{li2025automatic, juliani2026confounding}. Causal perspectives have been leveraged in \emph{policy space reduction} to eliminate reward-irrelevant actions and shrink the policy space \cite{zhang2020designing}, and in \emph{state space representation} \cite{wang2022causal, zhang2020invariant, wang2024building, gao2025bagged} to  aggregate similar states (a.k.a., state abstraction \citep{andre2002state, abel2018state,ni2024bridging}) and  learn latent representations \cite{lesort2018state, ota2020can}. 
All these methods assume a valid Markovian state representation is already given. 
Our work is complementary: we provide a procedure to construct a minimal Markovian state from raw observations using a longitudinal causal graph, and we propose MOSE to make it usable in deep RL. The resulting representation can serve as a foundation upon which existing causal RL techniques could be applied to potentially further improve performance.


\textbf{Redundancy and Exploration in Deep RL}\;\;
A separate line of work has shown that controlled redundancy via structured noise, data augmentation, or multi‑view inputs can improve learning and generalization in deep RL. Coarse representations may encourage the learner to ignore impertinent information \citep{sutton1998rl}. Noise‑based exploration methods (e.g., NoisyNet \citep{fortunato2018noisy} and curiosity‑driven exploration \citep{pathak2017curiosity}) inject stochasticity into the agent’s parameters or reward signal, smoothing the optimization landscape and preventing premature convergence. Deep RL agents have been observed to learn more efficiently when increasing the input dimensionality through high-dimensional representation learning \citep{ota2020can}. Data augmentation (e.g., random shifts, cropping) acts as an effective regularizer in visual RL, improving generalization \citep{laskin2020reinforcement,yarats2021image,stooke2021decoupling}. Multi‑view RL \citep{li2019multi,chen2021unsupervised,jangir2022look,hwang2023information} and contrastive representation learning \citep{sermanet2018time,laskin2020curl,stooke2021decoupling,laskin2022unsupervised,eysenbach2022contrastive} further demonstrate that providing multiple, partially redundant views of the same observation yields richer, more robust representations. Collectively, these findings support the intuition that deep networks benefit not from information‑theoretic minimalism, but from carefully designed redundancy. MOSE operationalizes this basic principle.

\vspace{-5pt}

\section{Preliminaries}\label{subsec:prelim}
\vspace{-5pt}
\textbf{Structural Causal Models \;} This work uses the formalisms of structural causal models (SCMs) \citep{pearl2000models} to bridge causal DAGs and MDPs. 
For explication, we begin by defining stationary SCMs for i.i.d. data and subsequently extend them to the time series setting.

\begin{definition}[Structural causal model (SCM), \citealt{bareinboim2022pch}] \label{def:scm}
    An SCM is a tuple $\mathcal{M} \coloneqq \langle \mathbf{U}, p(\mathbf{u}), \mathbf{X}, \mathcal{F} \rangle$ where $\mathbf{U} = \{U^i\}_{i=1}^n$ is the set of exogenous variables explained by  mechanisms external to $\mathcal{M}$, $p(\mathbf{u})$ is the distribution over $\mathbf{U}$, $\mathbf{X} = \{X^i\}_{i=1}^n$ is the set of endogenous variables explained by variables in $\mathbf{U} \cup \mathbf{X}$, and $\mathcal{F} = \{f^i\}_{i=1}^n$ is the set of structural functions such that $x^i = f^i(\mathbf{pa}_{X^i}, u^i)$ for endogenous parent set $\mathbf{pa}_{X^i}$ and exogenous context $u^i$.
\end{definition}
Given an SCM, $\mathcal{M}$ (Def.~\ref{def:scm}), its causal structure is encoded by a directed graph $\mathcal{G} = (\mathbf{X}, \mathbf{E})$ over variables $\mathbf{X}$, and $\mathbf{E}$ is the edge set. For a $\mathcal{G}$ that is linked with $\mathcal{M}$, an edge $X^i\to X^j$ iff $X^i\in\mathbf{pa}_{X^j}$.
We restrict our attention to 
\emph{directed acyclic graphs} (DAGs). We assume a positive SCM:  $p(x^i) > 0$ for every realization $X^i \in \mathbf{X} = x^i$ \citep{pearl_probabilities_1999}. We additionally assume causal sufficiency, i.e., the absence of latent confounding in $\mathcal{G}$. We assume a nonparametric SCM, but do not require jointly independent noise variables and are therefore not restricted to the \emph{non‑parametric structural equation model with independent errors} (NPSEM-IE) regime \citep{wang2025causal}.

\textbf{Extension to Time Series \;} This work considers the setting where SCM $\mathcal{M}$ represents a discrete-time multivariate stochastic process $\mathbf{X},\mathbf{R},\mathbf{A}$ over a finite horizon, so observations are temporally dependent and not i.i.d. The 
endogenous variables are 
\begin{align}
    \mathbf{X} \coloneqq \{\mathbf{X}_t\}_{t=0}^T =  \{\mathbf{X}_0, \dots, \mathbf{X}_T\}, \mathbf{R} \coloneqq \{R_t\}_{t=0}^T =  \{R_0, \dots, R_T\}, \mathbf{A} \coloneqq \{A_t\}_{t=0}^T =  \{A_0, \dots, A_T\},
\end{align}
where $\mathbf{X}_t \coloneqq \{X^i_t\}_{i=0}^{m-1}$, with $m$ variables at each time step, and $\mathbf{X}^i \coloneqq \{X^i_t\}_{t=0}^T$ for discrete time indices $t \in \mathbb{N}$. 
As in Definition \ref{def:scm}, $X^i_t$ is a function of its causal parents and an exogenous noise variable $U_t^i$. 
$\mathcal{G}$ is now a time series causal DAG over endogenous variables  $\{\mathbf{X}_t\}_{t=0}^T\cup\{R_t\}_{t=0}^T\cup\{A_t\}_{t=0}^T$. The graph containing every variable in the time series SCM is the \textit{full time graph}.

\begin{definition}[Full time graph, \citealt{runge2018causal,assaad2022survey,hasan2023survey}]
\label{def:full_time_graph}
    Let $\mathcal{G} = (\mathbf{X}\cup\mathbf{R}\cup\mathbf{A}, \mathbf{E})$ be a full time graph of multivariate process $\mathbf{X},\mathbf{R},\mathbf{A}$. Nodes are $X^i_t \in \mathbf{X}$, $R_t \in \mathbf{R}$, $A_t \in \mathbf{A}$ for all time $t \in [T]$. Edges $\mathbf{E}$ are lag-specific directed links $X^i_{t-k} \to X^j_t$ with a time lag of $k > 0$ for $i=j$ (\textit{autodependencies}) and $k \geq 0$ for $i \neq j$, and analogously for $\mathbf{R}$ and $\mathbf{A}$.
\end{definition}

We do not require \textit{causal stationarity} in $\mathcal{G}$ (also known as the \textit{consistency throughout time} assumption), which states that any link $X^i_{t-k} \to X^j_t$ with lag $k$ recurs for all $t$ \citep{runge2018causal}. 
Importantly, we assume the \textit{causal Markov condition}.

\begin{definition}[Causal Markov condition]
\label{def:markov_condition_dag}
    Let $\mathcal{\mathcal{G}} = (\mathbf{X} \cup \mathbf{A} \cup \mathbf{R},\mathbf{E})$ be a full time graph. For node $X^i_t \in \mathbf{X}$, let  $\mathbf{nd}_{X^i_t}$ denote its nondescendants and $\mathbf{pa}_{X^i_t}$ its parents. For every $X^i_t \in \mathbf{X}$, 
     \begin{align}
         X^i_t \ind \mathbf{nd}_{X^i_t} \mid \mathbf{pa}_{X^i_t}.
     \end{align}
\end{definition}
We do not make explicit assumptions over $\mathbf{A}, \mathbf{R}$.  Definition~\ref{def:markov_condition_dag} is important because with this, checking conditional independence between $X_{t_1}^{i_1}, X_{t_2}^{i_2}$ conditioned on $X_{t_3}^{i_3}$ is equivalent to check whether the $X_{t_3}^{i_3}$ blocks all path from $X_{t_1}^{i_1}$ to $X_{t_2}^{i_2}$ in the DAG $\mathcal{G}$ \citep{pearl_probabilities_1999}. 
Though the \textit{time series model with independent noise} (TiMINo) \citep{peters2013causal} is a popular extension of the NPSEM-IE for which causal Markov is known to hold, we find this setting overly restrictive for RL. In many RL applications like healthcare, the noise often has seasonal trends which breaks the i.i.d. assumption. Further, although there always exists an equivalent SCM describing the data when the underlying SCM is an MDP and the value space of $\mathbf{X}_t$ is countable (detailed discussion in Appendix~\ref{sec:valid-iid-noise}), this does not always hold when the space is not countable. Therefore, we relax the strong assumption of jointly independent noise to conditional independence as stated below.
\begin{assumption}[Conditional independence of exogenous variables]\label{assump:conditional-independence-noise}
    Given an SCM denoted by $\mathcal{M}$, we assume the exogenous variables $\mathbf{U}$ over $\mathbf{X}$ satisfy
$
        U^i_t \ind \left\{U_h^j:X_h^j\in\mathbf{nd}_{X_t^i}\right\} \mid \mathbf{pa}_{X^i_t}.$
\end{assumption}
Assumption~\ref{assump:conditional-independence-noise} states that all the influence of past noise to the current node $X_t^i$ are fully mediated through the parents of $X_t^i$. 
Assumption~\ref{assump:conditional-independence-noise}
ensures that the causal markov condition holds. Formally,
\begin{proposition}\label{th:causal-markov-condition}
    Under Assumption~\ref{assump:conditional-independence-noise}, for node $X^i_t \in \mathbf{X}$, let  $\mathbf{nd}_{X^i_t}$ denote its nondescendants and $\mathbf{pa}_{X^i_t}$ its parents. For every $X^i_t \in \mathbf{X}$, 
         $X^i_t \ind \mathbf{nd}_{X^i_t} \mid \mathbf{pa}_{X^i_t}$.
\end{proposition}
Proposition~\ref{th:causal-markov-condition} can be proven using the same technique as \cite{peters2013causal} and we provide proof in Appendix~\ref{sec:proof-causal-markov}. When Assumption~\ref{assump:conditional-independence-noise} fails, the causal Markov condition might also fail and it becomes unclear whether it is possible to recover the Markov state from the DAG alone. 
See Appendix~\ref{sec:dag-helpful-causal-markov} for discussion.
\begin{remark}[Relaxation of Assumption~\ref{assump:conditional-independence-noise}]
We only need \emph{causal markov condition} for Theorem~\ref{thm:criterion_valid}-\ref{thm:minimality}. Though Assumption~\ref{assump:conditional-independence-noise} is still a restrictive assumption to ensure causal markov condition, it already relaxes the independent noise assumption used in literature \citep{peters2013causal}. In this work, we focus on constructing markovian state from DAG under causal markov condition. Thus, finding the exact necessary and sufficient condition for causal markov condition is important future work.
\end{remark}

\textbf{W\textsuperscript{th} Order MDP \;} We consider an episodic finite-horizon problem with
horizon $T$. At each time step $t \in \{0,\dots,T\}$, the learner observes $m$ scalar variables, represented by 
$
\mathbf{X}_t = (X_{t}^0,\dots,X_{t}^m) \in \mathcal{X},
$
where $\mathcal{X}$ denotes the observation space. 
After choosing an action $A_t \in \mathcal{A}$, the learner receives a reward
$r_t$ that satisfies $\mathbb{E}\left[r_t\mid \mathbf{X}_t,A_t\right]=r(\mathbf{X}_t,A_t)$, where $r(\mathbf{X}_t,A_t)$ is the expected reward function.
We say the system is a $W^\text{th}$ order MDP if, after defining
the state
\begin{equation}\label{eq:Z_t}
Z_t = (\mathbf{X}_t,\mathbf{X}_{t-1},\dots,\mathbf{X}_{\max\{t-W,0\}}),
\end{equation}
the process $\{Z_t\}_{t=0}^T$
satisfies the Markov property, namely,
    \begin{equation}
\mathbb{P}(Z_{t+1} \mid  O_0,\dots,Z_t,  A_0,\dots, A_t)
=
\mathbb{P}(Z_{t+1} \mid Z_t, A_t).
\end{equation}

For any policy $\bm \pi = \{\pi_t\}_{t=0}^T$, we define the value function and $Q$-function
at time $t$ by
\begin{align*}
Q_t^{\bm \pi}(z_t,a)
\coloneqq
\mathbb{E}^{\bm \pi}\left[
\sum_{h=t}^T r(Z_h,A_h) \,|\, Z_t=z_t,\ A_t=a
\right], \;\;
V_t^{\bm\pi}(Z_t)
\coloneqq
\mathbb{E}^{\bm \pi}\left[
\sum_{h=t}^T r(Z_h,A_h) \,|\, Z_t=z_t
\right].
\end{align*}
By optimal Bellman equation \citep{sutton1998rl}, we have $     Q_t^*(z_t,a)= r(z_t,a)+\mathbb{E}\left[V_{t+1}^*(Z_{t+1})\mid Z_t=z_t,A_t=a\right],
 V_t^*(z_t)= \max_a Q_t^*(z_t,a),$
where $Q_t^*(z_t,a)=\max_{\bm\pi} Q_t^{\bm\pi}(z_t,a)$ and $V_t^*(z_t)=\max_{\bm\pi} V_t^{\bm\pi}(z_t)$. We further have the optimal policy $\bm\pi^*$ is given by $\pi^*_t(z_t)=\arg\max_aQ_t^*(z_t,a)$.

In the online setting, at each time $t$ of episode $k$, the learner observes $\mathbf{X}_t$ and then
chooses an action $A_t$ based on the history
$h_t = (\mathbf{X}_0,\dots,\mathbf{X}_t,\ A_0,\dots,A_{t-1},\ R_0,\dots,R_{t-1}).$
Then the learner observes a reward $r_t^{(k)}$.
Over $K$ episodes, the goal is to minimize the regret of online policy $\bm\pi^{(k)}$:
$\mathbb{E}\left[\sum_{k=1}^K V_0^*(z_0)-V_0^{\bm\pi^{(k)}}(z_0)\right].$
However, directly taking all variables in $Z_t$ as the state can substantially slow learning. In tabular RL, the regret scales polynomially with the number of states \citep{auer2008nearoptimal}, so when the state is
formed by stacking all $m$ coordinates across $W+1$ time steps, the effective state space can grow rapidly with both $m$ and $W$. 

We therefore seek a lower-dimensional state process that can preserve the Markov property and optimality. This leads to the following definition of a \textit{valid state}.

\begin{definition}[Valid States]
\label{def:valid_state} Consider a $W$\textsuperscript{th} order MDP, and let  $Z_t$ be the stacked observations from window $t-W$ to $t$, as defined in Eq. \eqref{eq:Z_t}. 
A sequence of random variable sets 
$\{S_t\}_{t=0}^T$ is called a \emph{valid state} if, for each $t$, $S_t$ is a subset of the observed variables contained in $\{\mathbf{X}_{i}\}_{i\in[t]}$ and the following two conditions hold:
\begin{enumerate}[leftmargin=*]
    \item \textbf{Markov Property.} For every $t$ and any sequence of actions, $\mathbb{P}\bigl(S_{t+1} \mid S_t, A_t, \dots, S_0, A_0\bigr) 
    = 
    \mathbb{P}\bigl(S_{t+1} \mid S_t, A_t\bigr).$
    \item \textbf{Optimality preservation.} For every $t$ and every realization $s_t\in Z_t, s_t\in S_t, z_t\in Z_t$,
$\arg\max_aQ^*_t\left(s_t,a\right)=\arg\max_aQ_t^*(z_t,a)$.
\end{enumerate}
\end{definition}
\vspace{-5pt}
\section{Minimal Markovian
State \& Multi-Order State Exposure}
\vspace{-5pt}
Section~\ref{sec:dag-mdp} presents the minimal Markovian state based on a known causal DAG.  Section~\ref{sec:mose} introduces \emph{Multi-Order State Exposure} (MOSE), which constructs the state that can be fed to \emph{any deep RL algorithm} as input when the DAG is unknown.

\subsection{From DAG to MDP: Minimal Markovian State Construction} \label{sec:dag-mdp}
We introduce a graphical criterion for constructing valid states   (Definition \ref{def:valid_state})
in Theorem \ref{thm:criterion_valid}, and then show that Algorithm \ref{alg:backward_state_construction}  constructs a minimal set satisfying this criterion (Theorem \ref{thm:minimality}, Lemma \ref{lem:parent_or_self}). Complete proofs are provided in Appendix \ref{sec:proofs}.


\begin{theorem}[Graphical criterion for valid state construction]
\label{thm:criterion_valid}
Let $\mathcal{G}$ be a full time graph as defined in Definition \ref{def:full_time_graph}. 
For each $t$, we are given $S_t \subset \cup_{j\in[t]} O_j$, which is a subset of state variables observed up to the current time. Let $\mathbf{A}_t :=  \{A_j\}_{j\in[t]}$ be the set of actions taken so far.
We say $\{S_t\}_{t\in[T]}$ are valid states (Definition~\ref{def:valid_state}) for all possible DAG if and only if the following conditions hold:  
\begin{enumerate}[leftmargin=15pt]
    \item \textbf{Reward parent inclusion.} \label{eq:reward-parent-include} For every $t$, the parents of the reward $R_t$ (as defined in the graph $\mathcal{G}$) are contained in ${S}_t$:
    $
    \mathbf{pa}_{R_t} \setminus \mathbf{A}_t  \subseteq {S}_t .$
    \item \textbf{Inclusion of extended next state parents.} \label{eq:state-parent-include} For every $t$, let $\mathcal{C}_{t+1}^*({S}_{t+1})$ be the closure of ${S}_{t+1}$ under same‑time parents, constructed iteratively as follows (by replacing ${S}_{t+1}$ with $Z$):
    \[
\mathcal{C}_{t+1}^0(Z) = Z,\;
\mathcal{C}_{t+1}^{k+1}(Z) = \mathcal{C}_{t+1}^{k}(Z) \;\cup\;
\Bigl( \bigcup_{Y \in \mathcal{C}_{t+1}^{k}(Z)} \bigl( \mathbf{pa}_Y \cap \mathbf{X}_{t+1} \bigr) \Bigr),\;
\mathcal{C}_{t+1}^*(Z) = \bigcup_{k\ge 0} \mathcal{C}_{t+1}^{k}(Z).
\]
    We require that every \emph{parent} of any $X \in \mathcal{C}_{t+1}^*({S}_{t+1}\setminus S_t)$
    that lies in the past (times $\le t$) and is not an action must be included in ${S}_t$:
$
\bigcup_{X \in \mathcal{C}_{t+1}^*({S}_{t+1}\setminus S_t)} \bigl( \mathbf{pa}_X \setminus (\mathbf{X}_{t+1}\cup\mathbf{A}_t) \bigr) \;\subseteq\; {S}_t .
$
    \item \textbf{Retension of Extended Neighbours} \label{eq:state-decendent-non} 
    For any $t$, let $N_{t+1}:=\{X_h^i: X_{h}^i\in S_{t+1}\setminus S_t,h\leq t\}$ represent the variables that is new in the next time state whose index is smaller than $t$. For any $s\leq t$, let $A_{t,s}$ to be the extended parents of variables indexed by $s$ that are in historic state, i.e. $A_{t,s}:=C_{s}^*\left(\bigcup_{h\leq t+1}S_{h}\cap \bm X_s\right)$. Then let $M_t$ be the moral graph for $\mathcal{G}$ constrained on variables $X_h^i$ in $A_{t,s}$ and $\textbf{pa}_{X_h^i}$ that are not in $\cup_{h\leq t}S_h$. For every maximal clique $K$ of $M_t$ such that $K\cap N_{t+1}\neq\emptyset$, the following must hold:
$
        \bigcup_{X_h^i\in A_{t,s},\\\{X_h^i\}\cup \textbf{pa}_{X_{h^i}}\cap K\neq\emptyset} \left(\{X_h^i\}\cup\textbf{pa}_{X_h^i}\right)\bigcap \left(\bigcup_{h\leq t} S_h\right)\subset S_t.
$
    
    \item \textbf{Persistence of included variables.}\label{eq:state-inclusion-time} If a variable appears in the state at two different times, it must also appear at all intermediate times:
    $
    \forall\, t_1 \le t_2 \le t_3,\;
    {S}_{t_1} \cap {S}_{t_3} \;\subseteq\; {S}_{t_2} .
    $
\end{enumerate}
\end{theorem}
Condition~\ref{eq:state-parent-include} ensures that $S_t$ captures every variable at times $\leq t$ that directly influences the extended next-state set $\mathcal{C}_{t+1}^*({S}_{t+1})$.
Any causal path from the earlier history to ($S_{t+1}$) must pass through one of these parents. Condition~\ref{eq:state-decendent-non} guarantees that markov blanket of $S_{t+1}$ when restricted on vertices indexed before $t$ is included in $S_t$.
Condition~\ref{eq:state-inclusion-time} 
ensures that when some variable is present in the state of two time steps, this will not cause the Markov property to fail because that variable will be in the state of every time step in between. 

In this work, we assume for simplicity, the action $A_t$ can only affect $\mathbf{X}_{t+1}$. When this assumption does not hold, we need to modify Theorem~\ref{thm:criterion_valid} to also include the parents of actions to the state to avoid edge cases like $X_{t-2}^i\to A_{t-2}\to X_t^i$.
We note that for a $W$\textsuperscript{th} order MDP, 
stacking all variables within the window length $W$
satisfies all four criteria and is a valid state.


However, a state construction $\{S_t\}_{t\in[T]}$
that satisfies the criteria under Theorem \ref{thm:criterion_valid} may contain redundant variables that are not needed to maintain validity. To remove redundant variables, we propose a backward construction procedure in Algorithm~\ref{alg:backward_state_construction}.
Algorithm~\ref{alg:backward_state_construction} constructs the state sets $\{S_t\}_{t\in[T]}$ backwards, starting from the empty set at time $T+1$.
At step $t$, we are given $S_{t+1}$  and an auxiliary set $\mathcal{C}_{t+1}$; together, $S_{t+1}\cup \mathcal{C}_{t+1}$ forms the extended next-state set $\mathcal{C}^*_{t+1}(S_{t+1})$ required by Condition~\eqref{eq:state-parent-include}.
We construct $S_t$ as follows. First, any variable in $S_{t+1}$ with time index $\leq t$ is included in $S_t$. This avoids recursively expanding their own parents, which would needlessly increase the state size. Second, to enforce Condition~\ref{eq:state-parent-include}, $S_t$ is augmented with all all parents of the extended next‑state set $S_{t+1}\cup \mathcal{C}_{t+1}$ that lie at times $\leq t$, together with parents of the reward $\mathbf{pa}_{R_t}$.
To illustrate the construction, we provide an example in Figure~\ref{fig:dag_mdp}.
In Theorem \ref{thm:minimality}, we prove that Algorithm \ref{alg:backward_state_construction} returns a \textit{minimal} set satisfying Theorem \ref{thm:criterion_valid} such that removal of any variable from $\mathbf{S}$ is impossible.

\begin{figure}[!t]
    \centering
    \resizebox{\columnwidth}{!}{%
\begin{tikzpicture}[scale=0.6, 
                    >={Stealth[width=4,length=4]},
                    every circle node/.style={draw,thick,scale=0.7},
                    background rectangle/.style={draw=platinum,fill=platinum,fill opacity=0.05,rounded corners=1ex},
                    show background rectangle,
                    line width=.6pt,
                    box/.style = {draw,very thick,CornflowerBlue,fill=CornflowerBlue,fill opacity=0.1,inner sep=7pt,rounded corners=3pt}]

    
    \node[circle] (w) at (-2,2) {$W_0$};
    \node[circle] (x) at (0,2) {$X_0$};
    \node[circle] (y) at (2,2) {$Y_0$};
    \node[circle] (z) at (-2,4) {$L_0$};
    \node[circle] (q) at (2,4) {$N_0$};
    \node[circle] (u) at (0,4) {$M_0$};
    \node[circle,WildStrawberry,fill=WildStrawberry!10] (r) at (0,0) {$R_0$};
    \node[circle,Blue,fill=Blue!10] (a) at (2,0) {$A_0$};

    \draw[->] (y) -- (r);
    \draw[->] (x) -- (r);
    \draw[->] (w) -- (r);
    \draw[->] (q) -- (y);
    \draw[->] (q) -- (x);
    \draw[->] (z) -- (w);
    \draw[->] (u) -- (x);
    \draw[->] (w) -- (x);
    \draw[->] (x) -- (y);

    \node[] (s0_label) at (-3.3,2) {\color{CornflowerBlue}$S_0$};
    \begin{pgfonlayer}{background}
        \fill[CornflowerBlue,opacity=0.3]
            (-2.8,4.6) --
            (-2.8,1.25) --
            (2.8,1.25) --
            (2.8,4.6) --
            (1.15,4.95) --
            (0,3.05) --
            (-1.15,4.95) -- cycle;
    \end{pgfonlayer}
    \draw[->,very thick,CornflowerBlue] (2,1.3) -- (a);


    \node[circle] (w_1) at (5,2) {$W_1$};
    \node[circle] (x_1) at (7,2) {$X_1$};
    \node[circle] (y_1) at (9,2) {$Y_1$};
    \node[circle] (z_1) at (5,4) {$L_1$};
    \node[circle] (q_1) at (9,4) {$N_1$};
    \node[circle] (u_1) at (7,4) {$M_1$};
    \node[circle,WildStrawberry,fill=WildStrawberry!10] (r_1) at (7,0) {$R_1$};
    \node[circle,Blue,fill=Blue!10] (a_1) at (9,0) {$A_1$};

    \draw[->] (y_1) -- (r_1);
    \draw[->] (x_1) -- (r_1);
    \draw[->] (w_1) -- (r_1);
    \draw[->] (q_1) -- (y_1);
    \draw[->] (q_1) -- (x_1);
    \draw[->] (z_1) -- (w_1);
    \draw[->] (u_1) -- (x_1);
    \draw[->] (y) -- (w_1);
    \draw[->] (w_1) -- (x_1);
    \draw[->] (x_1) -- (y_1);

    \node[] (s1_label) at (3.7,3.15) {\color{CornflowerBlue}$S_1$};
    \begin{pgfonlayer}{background}
        \fill[CornflowerBlue,opacity=0.3]
            (1.35,4.55) --
            (1.15,3.45) --
            (4.55,1.25) --
            (9.55,1.25) --
            (9.55,2.75) --
            (7.6,2.75) --
            (5.55,4.85) --
            (4.35,4.85) -- cycle;
    \end{pgfonlayer}
    \draw[->,very thick,CornflowerBlue] (9,1.3) -- (a_1);


    \node[circle] (w_2) at (12,2) {$W_2$};
    \node[circle] (x_2) at (14,2) {$X_2$};
    \node[circle] (y_2) at (16,2) {$Y_2$};
    \node[circle] (z_2) at (12,4) {$L_2$};
    \node[circle] (q_2) at (16,4) {$N_2$};
    \node[circle] (u_2) at (14,4) {$M_2$};
    \node[circle,WildStrawberry,fill=WildStrawberry!10] (r_2) at (14,0) {$R_2$};
    \node[circle,Blue,fill=Blue!10] (a_2) at (16,0) {$A_2$};

    \node[] (blank) at (18,2) {$\mathbf{\dots}$};

    \draw[->] (y_2) -- (r_2);
    \draw[->] (x_2) -- (r_2);
    \draw[->] (w_2) -- (r_2);
    \draw[->] (q_2) -- (y_2);
    \draw[->] (q_2) -- (x_2);
    \draw[->] (z_2) -- (w_2);
    \draw[->] (u_2) -- (x_2);
    \draw[->] (y_1) -- (w_2);
    \draw[->] (w_2) -- (x_2);
    \draw[->] (x_2) -- (y_2);
    \draw[->] (y_2) -- (blank);

    \node[] (s2_label) at (10.9,2.8) {\color{CornflowerBlue}$S_2$};
    \begin{pgfonlayer}{background}
        \fill[CornflowerBlue,opacity=0.3] \convexpath{y_2,x_2,w_2}{20pt};
    \end{pgfonlayer}
    \draw[->,very thick,CornflowerBlue] (16,1.3) -- (a_2);


    \draw[->] (z) edge[bend left=30] (z_1);
    \draw[->] (z_1) edge[bend left=30] (z_2);

    \draw[->] (q) edge[bend left=20] (q_2);

    \draw[->] (a) -- (4.457,1.55);
    \draw[->] (a_1) -- (11.457,1.55);

    \draw[->] (a) -- (r);
    \draw[->] (a_1) -- (r_1);
    \draw[->] (a_2) -- (r_2);
    \draw[->] (a_2) -- (blank);

    \node[] (label) at (18,0) {(\textbf{1})};
    
\end{tikzpicture}
\hspace{4mm}
\begin{tikzpicture}[scale=0.6, 
                    >={Stealth[width=4,length=4]},
                    every circle node/.style={draw,thick,scale=0.7},
                    background rectangle/.style={draw=platinum,fill=platinum,fill opacity=0.05,rounded corners=1ex},
                    show background rectangle,
                    line width=.6pt,
                    every edge quotes/.style={font=\footnotesize,fill=white,sloped}]

    \node[circle,CornflowerBlue,fill=CornflowerBlue!10] (s0) at (0,2.25) {$S_0$};
    \node[circle,CornflowerBlue,fill=CornflowerBlue!10] (s1) at (2,2.25) {$S_1$};
    \node[circle,CornflowerBlue,fill=CornflowerBlue!10] (s2) at (4,2.25) {$S_2$};

    \node[circle,Blue,fill=Blue!10] (a0) at (0,4.5) {$A_0$};
    \node[circle,Blue,fill=Blue!10] (a1) at (2,4.5) {$A_1$};
    \node[circle,Blue,fill=Blue!10] (a2) at (4,4.5) {$A_2$};

    \node[circle,WildStrawberry,fill=WildStrawberry!10] (y0) at (0,0) {$R_0$};
    \node[circle,WildStrawberry,fill=WildStrawberry!10] (y1) at (2,0) {$R_1$};
    \node[circle,WildStrawberry,fill=WildStrawberry!10] (y2) at (4,0) {$R_2$};

    \node[] (blank) at (6,2.25) {$\mathbf{\dots}$};
    \node[] (spacer) at (0,5.25) {};
    
    \draw[->] (s0) -- (s1);
    \draw[->] (s1) -- (s2);

    \draw[->] (s0) -- (a0);
    \draw[->] (s1) -- (a1);
    \draw[->] (s2) -- (a2);

    \draw[->] (s0) -- (y0);
    \draw[->] (s1) -- (y1);
    \draw[->] (s2) -- (y2);

    \draw[->] (a0) -- (s1);
    \draw[->] (a1) -- (s2);

    \draw[->] (a0) edge[bend left=30] (y0);
    \draw[->] (a1) edge[bend left=30] (y1);
    \draw[->] (a2) edge[bend left=30] (y2);

    \draw[->] (s2) -- (blank);
    \draw[->] (a2) -- (blank);

    \node[] (label) at (6,0) {(\textbf{2})};
\end{tikzpicture}%
}
\vspace{-10pt}
    \caption{From DAG to MDP. (\textbf{1}) Time-series causal DAG shown at time $t \in [0,2]$, where $S_0$, $S_1$, and $S_2$ are valid states selected by Algorithm \ref{alg:backward_state_construction}. (\textbf{2}) Causal DAG representation of the corresponding MDP \citep{bareinboim2025intro_causal_rl}, a subgraph of (\textbf{1}).}
    \label{fig:dag_mdp}
    \vspace{-\baselineskip}
\end{figure}

\begin{theorem}[Minimality of the state space returned by Algorithm~\ref{alg:backward_state_construction}]
\label{thm:minimality} Given a full time DAG $\mathcal{G}$,
let \(\{{S}_t\}_{t=0}^T\) be the state sequence returned by Algorithm~\ref{alg:backward_state_construction}. 
Then \(\{S_t\}_{t=0}^T\) is valid (satisfies Definition~\ref{def:valid_state}).
Moreover, it is \emph{minimal} in the following sense:
for any sequence \(\{\widetilde{S}_t\}_{t=0}^T\) such that
$
\widetilde{{S}}_t \subseteq {S}_t,\; \forall t,
\qquad\text{and}\qquad
\widetilde{{S}}_{t_0} \subsetneq {S}_{t_0} \;
\text{ for at least one time step } t_0.
$
There exists a transition probability model admissible by the
DAG such that $\{\widetilde{{S}}_t\}_{t=0}^T$ is invalid.
\end{theorem}

The proof of Theorem~\ref{thm:minimality} proceeds in two steps 1) verifying the criterion in Theorem~\ref{thm:criterion_valid} are satisfied naturally by construction and 2) to preserve optimality, the causal parents of the parents of reward has to be involved in the state as established by Lemma~\ref{lem:parent_or_self}.

\begin{wraptable}{r}{0.35\textwidth}
\vspace{-28pt}
    \centering
    \begin{tabular}{c|c}
        \hline
         &   Nodes Per State \\
        \hline
        $W=2$ & $[16.77,16.90]$  \\
        $W=5$ & $[28.06,28.41]$ \\
        $W=10$& $[35.17,35.93]$ \\
        \hline
    \end{tabular}
    \caption{Nodes per state (95\% CI).}
    \label{table:average-nodes}
    \vspace{-\baselineskip}
\end{wraptable}
To illustrate the scale of reduction of the state space size, we evaluate the number of variables  
contained in ${S}_t$ for randomly generated graphs. In Table~\ref{table:average-nodes} we report the  95\% confidence interval for the average number of variables contained over all horizons. The confidence interval is calculated based on 100 randomly generated graphs with 10 nodes per time step. The exact generation process is the same as the one shown in Section~\ref{sec:experiment}. This illustrates that we can often scale down the state space size relative to the full window state.  
\vspace{-5pt}
\subsection{Multi-Order State Exposure}\label{sec:mose}
\vspace{-5pt}
Deep RL is known to suffer from decreasing expressivity and generalizability \citep{sokar2023dormant} due to overfitting to early samples \citep{cobbe2019quantifying}. Feeding a neural network the minimal Markovian representation can further worsen this behavior, as it removes many variables from observation that can provide additional information to  help the network adapt to new samples. This is consistent with our observations in Figure~\ref{fig:ANM-uniform}, where the minimal Markovian representation did not improve over window policies when $W=5$. To confirm this, we plot the difference between training and testing loss when the state is minimal Markovian state. The curve is averaged over $H$ horizons using linear ANM detailed in Appendix~\ref{app:DGP_details}. As shown in Figure~\ref{fig:loss_curve_30_5_10}, the DAG state starts with good generalization in the beginning and fluctuates toward negative as more samples come in. This indicates that the model using DAG state overfits to early samples and cannot adjust to later samples, as the testing loss consistently becomes larger. Meanwhile, though the magnitude of stacked window state is smaller than DAG state, it still fluctuates. In contrast, MOSE and Causal MOSE, which will be introduced later, consistently have near-zero loss difference, showing the benefits of controlled redundancy. We include the per-horizon loss curve in Appendix~\ref{sec:per-horizon-loss-difference}. 

\begin{wrapfigure}{r}{0.35\textwidth}
     \vspace{-25pt}
        \centering
        \includegraphics[width=\linewidth]{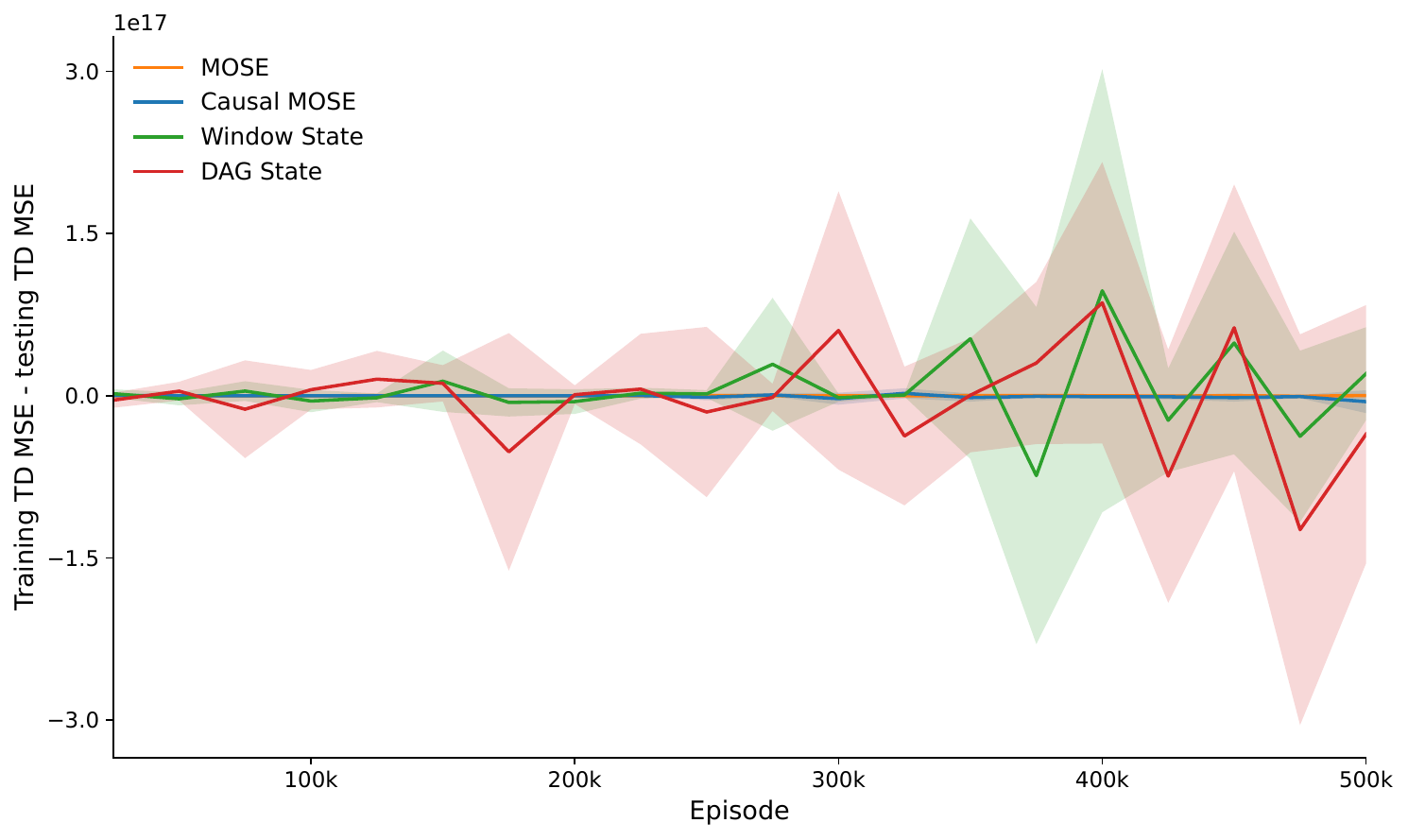}
        \vspace{-15pt}
        \caption{Loss Difference Averaged over Horizons.}
        \label{fig:loss_curve_30_5_10}
            \vspace{-\baselineskip}   
\end{wrapfigure}
To address this issue, we introduce \textit{Multi-Order State Exposure} (MOSE; Algorithm~\ref{alg:mose}). Based on the idea that selecting features jointly with the model that uses them will improve model performance \citep{sutton2019bitter,balin2019concrete}, MOSE explicitly adds a selection layer before the final MLP that estimates the value function. As shown in Figure~\ref{figure:mose-architecture}, MOSE operates on a history that is either the raw observation or the belief history learned by a sequential architecture like LSTM. We first embed each dimension into a higher-dimensional space by $E_t^i=\text{tanh}\left(W^iX_{t}^i+b_t^i\right)$. We use a separate weight $W$ for different dimensions of $X$ \citep{gorishniy2021revisiting} because each dimension of $X$ describes a different aspect of the environment. We let the same dimension of $X$ across different time steps share the same weight because they should lie in the same direction, with the effect of time introduced by a bias term. To choose which $X_h^i$ in the history $Z_t$ should be part of the state, we need to calculate the importance of this dimension to estimating value functions. Attention provides a natural way to parameterize this dependence on the value-estimation task \citep{vaswani2017attention}. Accordingly, we introduce a single learned query $q_0$, shared across candidate coordinates and trained jointly with the Q-network. Its association with value estimation comes from the learning objective: gradients from value prediction determine which information the query learns to retrieve. 

The selector uses this query to obtain a context for evaluating the current history. Let $i$ index the available variable--lag pairs, i.e., $i=(h,j)$ for $t-W\leq h\leq t, 0\leq j\leq m$, and let $k_{t,i},v_{t,i}\in\mathbb{R}^{d}$ denote the key and value obtained from the embedding of coordinate $i$ at time $t$. The attention read is
\begin{equation}
    \alpha_{t,i}
    =
    \frac{\exp\!\left(q_0^\top k_{t,i}/\sqrt{d}\right)}
         {\sum_j \exp\!\left(q_0^\top k_{t,j}/\sqrt{d}\right)},
    \qquad
    c_t = \sum_i \alpha_{t,i}v_{t,i}.
\end{equation}
 We then use $c_t$ to score the same candidate coordinates by $r_{t,i} = \frac{c_t^\top k_{t,i}}{\sqrt{d}}$. This second step allows selection to reflect the information retrieved from the current history while retaining a single independently learned query. The selection layer converts these scores by comparing the sigmoid of score and a learnable threshold $\tau$. We learn $\tau$ using a straight-through estimator \citep{bengio2013estimating} with hard gates $\mathbf{1}\{\sigma(z)\geq\tau\}$ in the forward pass and the differentiable surrogate $\sigma(\sigma(z)-\tau)$ in the backward pass. When $\sigma(r_{t,i})\geq \tau$, then the corresponding $X_{h}^j$ will be retained. Otherwise the value will be replaced by 0.
The Q-network receives the selected original observations together with their mask. Preserving the observations retains their numerical values, while including the mask distinguishes an excluded coordinate from an observed zero. This mask can be seen as an artificial missingness because the corresponding element is not helping the value function estimation. Therefore, we also include the mask as input to the value function following the common structure when dealing with missingness in MLP \citep{le2020neumiss}.

We initialize the threshold so that all available coordinates are selected. The selector and Q-network are then trained jointly using the temporal-difference objective. Starting with full selection gives the network access to potentially useful redundancy before learning which inputs it can dispense with. Because the gates remain trainable and the selector continues to inspect the available history, these decisions can be revised as learning proceeds. The aim is thus to learn a compact input representation suited to the Q-network, with room to retain redundant coordinates when they make value estimation easier. We have done ablations to replace the attention mechanism with a simple weight to directly represent the score and the results are deferred to Appendix~\ref{sec:ablation-attention}. When DAG information is available, we use the minimal Markovian state to specify inputs that MOSE must retain. We initialize the corresponding input gates to $1$ and keep them fixed at $1$ throughout training. The remaining gates are learned as before. Therefore the DAG state determines which inputs must remain available, and end-to-end selection determines which additional inputs are useful. To further emphasize DAG state, we additionally add an gradient update on the Q network using only DAG state as input (Algorithm~\ref{alg:causal-mose}).

\begin{figure}[!t]
\label{figure:mose-architecture}
\centering
\begingroup
\definecolor{moseInk}{HTML}{333333}
\definecolor{moseBlue}{HTML}{4B8DC5}
\definecolor{moseRed}{HTML}{D96867}
\definecolor{moseQuery}{HTML}{AD99CF}
\definecolor{moseContext}{HTML}{80B8CA}
\definecolor{moseGray}{HTML}{DADADA}
\def\moseScalarWidth{1.06}
\def\moseVectorWidth{1.90}
\def\moseCellHeight{.50}
\tikzset{
  moseArrow/.style={-{Latex[length=2mm,width=1.45mm]},draw=moseInk,line width=.8pt},
  moseFine/.style={-{Latex[length=1.8mm,width=1.25mm]},draw=moseInk!75,line width=.65pt},
  moseTitle/.style={font=\large\bfseries,align=center,text=moseInk},
  moseLabel/.style={font=\normalsize,align=center,text=moseInk},
  moseCell/.style={draw=moseInk,line width=.7pt,minimum width=\moseScalarWidth cm,
    minimum height=\moseCellHeight cm,inner sep=0pt,outer sep=0pt,align=center,font=\small},
  mosePanel/.style={draw=moseInk!50,line width=.65pt},
}
\newcommand{\moseVector}[5]{%
  \path[draw=moseInk!80,line width=.5pt,fill=#4,fill opacity=#5]
    ({#2-\moseVectorWidth/2},{#3-\moseCellHeight/2})
    rectangle ({#2+\moseVectorWidth/2},{#3+\moseCellHeight/2});
  \coordinate (#1-west) at ({#2-\moseVectorWidth/2},#3);
  \coordinate (#1-east) at ({#2+\moseVectorWidth/2},#3);
  \coordinate (#1-north) at (#2,{#3+\moseCellHeight/2});
  \coordinate (#1-south) at (#2,{#3-\moseCellHeight/2});
}
\hyphenpenalty=10000\exhyphenpenalty=10000
\resizebox{\linewidth}{!}{%
\begin{tikzpicture}[x=1cm,y=1cm,text=moseInk,font=\large]

\node[moseTitle] at (0,7.35) {(Belief) History};
\node[moseTitle] at (3.5,7.35) {Embedding};
\node[moseTitle] at (9.15,7.35) {Tokens};
\node[moseLabel,font=\Large] at (9.15,6.35) {$E$};
\path[mosePanel] (7.35,-.47) rectangle (10.33,5.97);
\foreach \tm/\mid/\top/\bot in {t/4.65/5.60/3.70,{t-W}/.75/1.70/-.20}{
  \node[moseLabel] at (0,{\top+.62}) {$\mathbf{X}_{\tm}$};
  \node[moseLabel] at (2.15,{\top+.62}) {$W$};
  \node[moseLabel] at (5.00,{\top+.62}) {$b$};
  \node[font=\Large,anchor=east] at (-1.02,\mid) {$\tanh$};
  \draw[draw=moseInk,line width=1pt] (-.62,{\top+.13}) -- (-.80,{\top+.13})
    -- (-.80,{\bot-.13}) -- (-.62,{\bot-.13});
  \draw[draw=moseInk,line width=1pt] (6.06,{\top+.13}) -- (6.24,{\top+.13})
    -- (6.24,{\bot-.13}) -- (6.06,{\bot-.13});
  \path[draw=moseInk!70,line width=.6pt,fill=moseGray!12]
    (-.53,\bot) rectangle (.53,\top);
  \path[draw=moseInk!70,line width=.6pt,fill=moseGray!12]
    (1.20,\bot) rectangle (3.10,\top);
  \path[draw=moseInk!70,line width=.6pt,fill=moseGray!12]
    (4.05,\bot) rectangle (5.95,\top);
  \path[draw=moseInk!70,line width=.6pt,fill=moseGray!12]
    (8.20,\bot) rectangle (10.10,\top);
  \node[font=\Large] at (.91,\mid) {$\times$};
  \node[font=\Large] at (3.58,\mid) {$+$};
  \foreach \xx in {0,2.15,5.00,9.15}
    \node[font=\small] at (\xx,\mid) {$\vdots$};
  \draw[draw=moseInk,line width=.8pt]
    (8.05,\top) .. controls (7.84,\top) and (7.84,{\top-.15}) .. (7.84,{\top-.30})
    -- (7.84,{\mid+.22}) .. controls (7.84,{\mid+.07}) and (7.72,{\mid+.07}) .. (7.68,\mid)
    .. controls (7.72,{\mid-.07}) and (7.84,{\mid-.07}) .. (7.84,{\mid-.22})
    -- (7.84,{\bot+.30}) .. controls (7.84,{\bot+.15}) and (7.84,\bot) .. (8.05,\bot);
  \node[font=\normalsize] at (7.53,\mid) {$m$};
}
\foreach \id/\tm/\dim/\yy/\col/\alpha in {
  a/t/1/5.35/moseBlue/.42,b/t/m/3.95/moseBlue/.78,
  c/{t-W}/1/1.45/moseRed/.42,d/{t-W}/m/.05/moseRed/.78}{
  \node[moseCell,fill=\col,fill opacity=\alpha,text opacity=1]
    (hist-\id) at (0,\yy) {$X_{\tm}^{\dim}$};
 \moseVector{weight-\id}{2.15}{\yy}{\col}{.65}
  \moseVector{bias-\id}{5.00}{\yy}{\col}{\alpha}
  \moseVector{tok-\id}{9.15}{\yy}{\col}{\alpha}
}
\foreach \id/\yy in {a/5.35,b/3.95,c/1.45,d/.05}
\draw[moseArrow] (6.24,\yy) -- (tok-\id-west);
\foreach \xx in {0,2.15,5.00}
  \node at (\xx,3.02) {$\vdots$};
\node at (9.15,2.70) {$\vdots$};
\draw[draw=moseInk,line width=.8pt]
  (8.20,-.67) .. controls (8.20,-.85) and (8.32,-.85) .. (8.50,-.85)
  -- (8.93,-.85) .. controls (9.08,-.85) and (9.08,-.97) .. (9.15,-1.01)
  .. controls (9.22,-.97) and (9.22,-.85) .. (9.37,-.85)
  -- (9.80,-.85) .. controls (9.98,-.85) and (10.10,-.85) .. (10.10,-.67);
\node[font=\normalsize] at (9.15,-1.30) {$d$};

\begin{scope}[xshift=8cm]
\node[moseTitle] at (8.4,7.35) {Attention};
\node[moseTitle] at (12.85,7.35) {Context};
\node[moseTitle] at (16.60,7.35) {Scores};
\node[moseTitle] at (21.90,7.35) {Mask};
\node[moseTitle] at (26.65,7.35) {Constructed\\state};
\node[moseTitle] at (30.10,7.35) {MLP};
\path[mosePanel] (2.80,-.86) rectangle (11.65,6.65);
\node[moseLabel] at (5.60,6.25) {Keys $K$, values $V$};
\node[moseLabel] at (15.75,6.25) {Same keys $K$};
\path[mosePanel] (14.18,-.86) rectangle (18.45,6.65);

\moseVector{seed}{9.80}{5.60}{moseQuery}{.65}
\node[moseLabel] at (9.80,6.20) {Learnable query $q_0$};
\node[font=\normalsize,inner sep=2pt,align=center] (readout) at (9.80,2.77)
  {$\displaystyle\operatorname{softmax}\!\left(\frac{q_0K^{\mathsf T}}{\sqrt d}\right)V$};
\draw[moseArrow,draw=moseQuery!85!black] (seed-south) -- (readout.north);
\draw[draw=moseInk!60,line width=.65pt]
  (7.65,5.91) -- (7.87,5.91) -- (7.87,-.51) -- (7.65,-.51);
\draw[moseFine] (7.87,2.77) -- (readout.west);

\moseVector{context}{12.825}{2.77}{moseContext}{.65}
\node[moseLabel] at (12.825,3.39) {$c$};
\draw[moseArrow] (readout.east) -- (context-west);
\draw[moseFine,draw=moseContext!85!black] (context-east) -- (14.52,2.77);
\draw[draw=moseContext!85!black,line width=.8pt] (14.52,.05) -- (14.52,5.35);

\foreach \id/\tm/\dim/\yy/\col/\alpha/\bit in {
  a/t/1/5.35/moseBlue/.42/1,b/t/m/3.95/moseBlue/.78/0,
  c/{t-W}/1/1.45/moseRed/.42/1,d/{t-W}/m/.05/moseRed/.78/0}{
  \moseVector{key-\id}{5.70}{\yy+.32}{\col}{\alpha}
  \moseVector{val-\id}{5.70}{\yy-.32}{\col}{\alpha}
  \draw[moseFine] (tok-\id-east) -- (3.25,\yy)
    -- node[midway,above=2pt,font=\small,inner sep=1pt] {$\times W_K$} (key-\id-west);
  \draw[moseFine] (3.25,\yy)
    -- node[midway,below=2pt,font=\small,inner sep=1pt] {$\times W_V$} (val-\id-west);
  \fill[moseInk!75] (3.25,\yy) circle (.85pt);
  \node[anchor=west,font=\normalsize,inner sep=1.5pt] at (6.73,{\yy+.32}) {$k_{\tm}^{\dim}$};
  \node[anchor=west,font=\normalsize,inner sep=1.5pt] at (6.73,{\yy-.32}) {$v_{\tm}^{\dim}$};

  \moseVector{again-\id}{15.75}{\yy}{\col}{\alpha}
  \node[font=\normalsize] at (15.75,{\yy+.42}) {$k_{\tm}^{\dim}$};
  \draw[moseFine,draw=moseContext!85!black] (14.52,\yy) -- (again-\id-west);
  \fill[moseContext!85!black] (14.52,\yy) circle (.95pt);
  \node[moseCell,fill=\col,fill opacity=\alpha,text opacity=1]
    (score-\id) at (17.58,\yy) {$s_{\tm}^{\dim}$};
  \draw[moseFine] (again-\id-east) -- (score-\id.west);

  \ifnum\bit=1
    \def\maskfill{\col}\def\maskopacity{\alpha}\def\selectedvalue{$X_{\tm}^{\dim}$}
  \else
    \def\maskfill{moseGray}\def\maskopacity{.35}\def\selectedvalue{$0$}
  \fi
  \node[moseCell,fill=\maskfill,fill opacity=\maskopacity,text opacity=1] (mask-\id) at (24.10,\yy) {$\bit$};
  \draw[moseFine] (23.15,\yy) -- (mask-\id.west);
  \node[moseCell,fill=\maskfill,fill opacity=\maskopacity,text opacity=1] at (26.08,\yy) {\selectedvalue};
  \node[moseCell,fill=\maskfill,fill opacity=\maskopacity,text opacity=1] at (27.40,\yy) {$\bit$};
}
\foreach \yy in {4.65,.75}{
  \foreach \xx in {17.58,24.10,26.08,27.40}
    \node[font=\small] at (\xx,\yy) {$\vdots$};
}
\foreach \yy in {4.65,.75}
  \node[inner sep=0pt] at (5.70,\yy) {\scalebox{.55}{$\vdots$}};
\foreach \yy in {4.86,.96}
  \node[font=\small] at (15.75,\yy) {$\vdots$};
\foreach \xx in {5.70,15.75,17.58,24.10,26.08,27.40}
  \node at (\xx,2.70) {$\vdots$};
\node[font=\normalsize] at (16.32,-.58)
  {$s_h^i=c(k_h^i)^{\mathsf T}/\sqrt d$};

\draw[draw=moseInk!60,line width=.65pt]
  (18.70,5.62) -- (18.88,5.62) -- (18.88,-.22) -- (18.70,-.22);
\node[draw=moseInk,line width=.7pt,fill=moseGray!18,
  inner xsep=7pt,inner ysep=10pt,font=\large] (maskrule) at (21.00,2.77)
  {$\mathbf{m}=\mathbf{1}\{\sigma(s)\ge\tau\}$};
\draw[moseFine] (18.88,2.77) -- (maskrule.west);
\node[moseLabel,font=\normalsize] at (21.00,1.78) {Learnable threshold $\tau$};
\draw[moseFine] (maskrule.east) -- (23.15,2.77);
\draw[draw=moseInk!75,line width=.65pt] (23.15,.05) -- (23.15,5.35);
\path[mosePanel] (23.45,-.47) rectangle (24.75,5.97);

\path[mosePanel] (25.15,-.86) rectangle (28.05,6.65);
\node[font=\normalsize] at (26.08,6.25) {$Z_t\odot\mathbf{m}$};
\node[font=\normalsize] at (27.40,6.25) {$\mathbf{m}$};
\node[font=\normalsize] at (26.60,-.46) {$[Z_t\odot\mathbf{m};\,\mathbf{m}]$};
\draw[moseArrow] (24.75,2.77) -- (25.15,2.77);
\foreach \xx/\hh/\shade in {29.20/3.90/25,29.85/3.0/38,30.50/2.10/52}
  \path[draw=moseInk,line width=.7pt,fill=moseQuery!\shade!white]
    ({\xx-.14},{2.77-\hh/2}) rectangle ({\xx+.14},{2.77+\hh/2});
\draw[moseArrow] (28.05,2.77) -- (29.06,2.77);
\draw[moseFine] (29.34,2.77) -- (29.71,2.77);
\draw[moseFine] (29.99,2.77) -- (30.36,2.77);
\end{scope}
\draw[moseArrow] (-2.70,5.35) -- (-2.70,-2.25)
  -- (34.60,-2.25) -- (34.60,-.86);
\draw[draw=moseInk,line width=.65pt] (hist-a.north) -- (0,5.97) -- (-2.70,5.97) -- (-2.70,5.35);
\draw[draw=moseInk,line width=.65pt] (hist-c.north) -- (0,2.08) -- (-2.70,2.08);
\node[font=\normalsize,fill=white,inner sep=3pt] at (20.20,-2.25)
  {Original history $Z_t=(\mathbf{X}_t,\ldots,\mathbf{X}_{t-W})$};
\end{tikzpicture}%
}
\endgroup
\vspace{-10pt}
\caption{\textbf{MOSE architecture.} A learned query $q_0$ attends to embedded history coordinates to form a context for assessing their relevance to Q-value estimation. This context scores each coordinate through its key to determine a selection mask. The masked observations and mask form the state passed to the MLP Q-network.}
\vspace{-\baselineskip}
\end{figure}
\vspace{-10pt}
\section{Experiments}\label{sec:experiment}
\vspace{-10pt}


We tested the benefits of MOSE (Algorithm~\ref{alg:mose}) and Causal MOSE (Algorithm~\ref{alg:causal-mose}) compared with naively stacking Window State: Defining all observations within a time window $w$ as state \citep{mnih2015human,hessel2018rainbow} and DAG-State: minimal valid state representation constructed by Algorithm~\ref{alg:backward_state_construction}. We evaluate the performance on synthetic data generating processes (DGPs) and Atari games \citep{10.5555/3304652.3304802}. We use a variant of Double Q Learning \citep{vanhasselt2015deepreinforcementlearningdouble} as the RL algorithm for synthetic environments and R2D2 \citep{kapturowski2018recurrent} for Atari games.

\textbf{Synthetic DGPs \;}  
We first randomly generate a graph with 10 nodes per time step and the causal parents of each node $X_t^i$ can only lie in $\{t-W,t-1,t\}$
where $W$ denotes the order of MDP and can take value in $\{5,10\}$ in our experiment. 
%
Given the causal graph,  data were sampled from additive noise models (ANMs) \citep{hoyer2008nonlinear,peters2014causal}.
We describe the details of the DGPs in Appendix~\ref{app:DGP_details}.
We compare two length of horizon per episode as 25 and 30 and we feed our state to a variant of Double Q Learning \citep{vanhasselt2015deepreinforcementlearningdouble} for episodic finite RL. 
Linear ANM \eqref{eq:linear-anm} are included in main body and
nonlinear ANM results are deferred to Appendix~\ref{appendix:experimental_details}.

\textbf{Atari Games \;}
We additionally evaluate MOSE
on the Atari game. In this environment, episode lengths are policy-dependent, since an episode terminates once the agent loses all lives. For this reason, we formulate the problem as an infinite-horizon discounted $W^{th}$ order stationary MDP; the formal definition is deferred to Appendix~\ref{appendix:prelims}. As the underlying RL backbone, we use R2D2~\citep{kapturowski2018recurrent}, a strong value-based deep RL algorithm. We compare MOSE against vanilla R2D2, where vanilla R2D2 directly uses the hidden states learned by LSTM as its input state. We claim that hidden states at current time might not be sufficient and it is beneficial to choose which dimensions of hidden states are enough. Therefore, we use MOSE to operate on the belief history learned by LSTM. Since the ground truth causal graph of the game environment is unavailable, we do not include Causal-MOSE in this experiment. For both methods, we train the agent for $2$ million environment interactions. Every $50{,}000$ environment steps, we evaluate the current policy over $10$ episodes and report the average episodic reward across these evaluation episodes. 


\vspace{-10pt}
\subsection{Experimental Results}
\vspace{-10pt}
\begin{figure}[t]
    \centering
    \includegraphics[width=\linewidth]{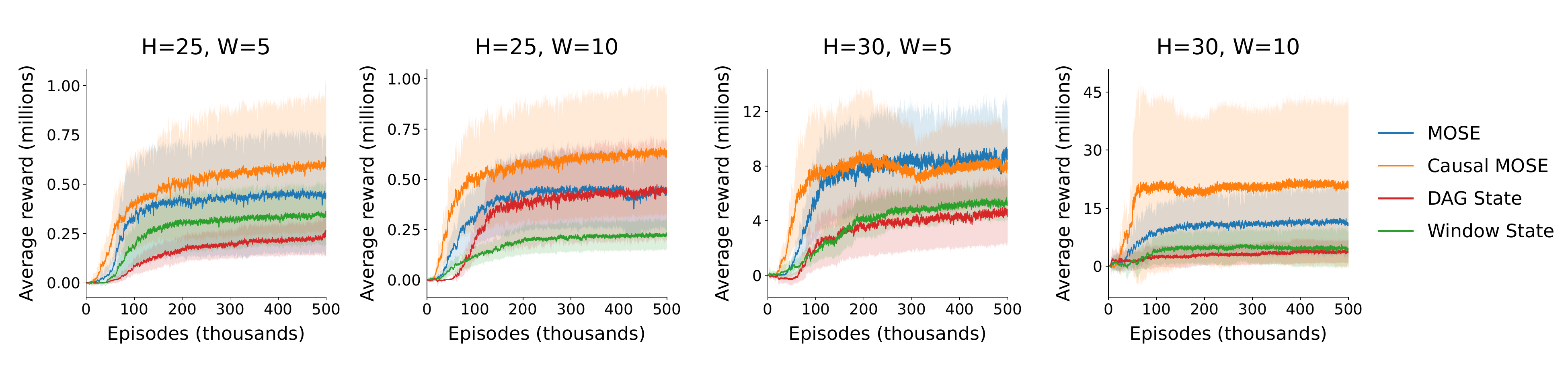}
    \vspace{-20pt}
    \caption{Average per-episode reward averaged over 18 instances. From left to right: $5^{\text{th}}$-order MDP, Horizon 25, $10^{\text{th}}$-order MDP, Horizon 25, $5^{\text{th}}$-order MDP, Horizon 30,$10^{\text{th}}$-order MDP, Horizon 25.  Linear ANM (Eq.~\eqref{eq:linear-anm})
    }
    \label{fig:ANM-uniform}
    \vspace{-\baselineskip}
\end{figure}
\textbf{Results Using Eq.~\eqref{eq:linear-anm} \;}
%
%
Figure~\ref{fig:ANM-uniform} shows a clear and consistent pattern across all three choices of the order $W$: MOSE outperforms all competing benchmarks throughout the full training horizon of 500K episodes for Linear ANM. The fact that this advantage appears for every tested $W$ and $H$ suggests that the gain from MOSE is not tied to a particular order and horizon length.
By selectively constructing the state representation, MOSE reduces the burden of learning from irrelevant or weakly informative variables for $Q$-function, leading to better performance across training.

A particularly notable observation from Figure~\ref{fig:ANM-uniform} is that the benchmark based only on the minimal Markovian state does \emph{not} improve performance, and in fact remains worse to window policies across 3 $(W,H)$ of 4 experiments. This is an important finding. From a theoretical perspective, the minimal Markovian state is sufficient to preserve the optimal control problem. However, the figure suggests that such minimality alone does not necessarily translate into the best empirical performance when the value function is approximated by a neural network. This indicates that although some variables are redundant for satisfying the Markov property, they may still contain auxiliary statistical information that helps the network estimate the $Q$-function more accurately in finite samples. Removing all such variables may therefore make the representation too sparse from the standpoint of function approximation, even if it remains sufficient in principle for optimal control.

This interpretation is further supported by the comparison between MOSE and Causal MOSE in Figure~\ref{fig:ANM-uniform}. When the minimal Markovian state is provided as prior structural information, Causal MOSE consistently outperforms MOSE over the entire 500K-episode training period. This shows that by fixing DAG state to be always input to the model, Causal MOSE can selectively choose which redundant information is needed for statistical efficiency and build state around DAG state.
Therefore, the best performance is achieved by combining a principled Markovian core with a mechanism that can incorporate extra informative variables. 

Finally, we plot the false negative and positive rate of the state constructed by MOSE compared with DAG state to check what state MOSE is trying to construct. As we can see from Figure~\ref{fig:false-positive-anm} the false positive rate of both MOSE and Causal MOSE is dropping and Causal MOSE dropped faster than MOSE. This shows that MOSE are slowly learning the DAG state and with DAG state hard encoded, Causal MOSE can drop truly redundant information more quickly. Though the false negative rate of MOSE is increasing, this is due to under finite horizon, MOSE may falsely drop important information and need more samples to collect back. Detailed discussions are in Appendix~\ref{sec:mose-dag}.


\textbf{Results on Atari Games \;}
Table~\ref{tab:atari-final-scores} shows that MOSE outperforms vanilla R2D2 after $2$M training samples, improving mean scores by \textbf{25.8\% on BattleZone} and \textbf{34.7\% on Krull}. Specific learning curves are deferred to Appendix~\ref{sec:learning-curve-atari}. 
\begin{wraptable}{l}{0.6\textwidth}
\vspace{-10pt}
    \centering
    \footnotesize
    \setlength{\tabcolsep}{4pt}
    \begin{tabular}{lcc}
        \toprule
        Game & R2D2 & MOSE \\
        \midrule
        BattleZone & $12{,}435.0 \pm 1{,}009.4$
                   & $15{,}640.0 \pm 5{,}409.6$ (\bf+25.8\%) \\
        Krull      & $5{,}556.9 \pm 888.6$
                   & $7{,}483.2 \pm 77.7$ (\bf+34.7\%) \\
        \bottomrule
    \end{tabular}
    \caption{Final Average results over past 10 evaluations after 2M training samples.}
    \label{tab:atari-final-scores}
\end{wraptable}
A possible explanation is that learned belief states are not necessarily Markovian: the LSTM may forget information relevant to control, making past belief states useful for current $Q$-value estimation. MOSE selects relevant components across the belief-state history, providing access to this information while avoiding the high-dimensional input produced by concatenating full past belief states. 
Furthermore, dimension-wise selection allows gradients to propagate back to the LSTM, guiding it toward learning belief representations that better retain control-relevant information.
\section*{AI use statement}
In this work, we used generative AI tools for 1) Generating synthetic data sets; 2) implementing methods.
We have not used generative AI tools for 1) helping develop theoretical models or conceptual frameworks; 2) formulating mathematical claims; 3) providing critical ingredients for proving mathematical claims and assisting in the writing of proofs; 4) proposing or refine hypotheses; 5) designing or provide feedback on research  methodology or experiments; 6)  supporting qualitative and thematic data analysis; 7) interpreting results
and assist with translation, clean and reformat dataset are not applicable to this work.

We have reviewed all AI-assisted work. LLM-generated code was verified and tested for correctness. We take responsibility for the final content of this work, including text, claims or artifacts produced with the aid of generative AI.
\clearpage
\bibliography{references}
\bibliographystyle{iclr2027_conference}


\clearpage

\appendix
\setcounter{figure}{0} 
\setcounter{table}{0}
\setcounter{algorithm}{0}
\renewcommand\thefigure{\thesection.\arabic{figure}}  
\renewcommand\thetable{\thesection.\arabic{table}}  
\renewcommand\thealgorithm{\thesection.\arabic{algorithm}}

\addcontentsline{toc}{section}{Appendix} 
\part{Appendix} 
\parttoc 
\section{Limitations}
In this work, we established a formal construction of minimal markovian state and design MOSE and Causal MOSE that can leverage the minimality of the state we constructed in deep RL. Several limitations remain. First, our approach assumes a known full time graph; learning such a DAG from high-dimensional longitudinal data, particularly with temporally correlated noise, remains an open challenge. Though state constructed by MOSE may be converging to minimal markovian state, its convergence speed has not been established and would be interesting future work. Second, integrating our framework with other causal RL techniques could further improve performance. These would be interesting future research directions.
\section{Extended Related Works}
\label{appendix:related_works}

\textbf{Longitudinal Causal Discovery}\;\; Despite extensive research addressing causal discovery in time series \citep{assaad2022survey,hasan2023survey}, longitudinal causal discovery in RL remains an open problem. Existing  longitudinal discovery approaches include Granger causality \citep{granger1969investigating,geweke1982measurement}, neural Granger causality \citep{marcinkevivcs2021interpretable,tank2022neural,lowe2022amortized}, methods that assume functional causal models \citep{fujiwara2023causal},  constraint-based algorithms \citep{entner2010causal,runge2019detecting,pmlr-v124-runge20a,gunther2023causal},  
and other deep learning approaches \citep{pamfil2020dynotears,bellot2022neural,sun2023nts}. 
These methods typically make   stationarity assumptions and sometimes deploy classical CITs that assume i.i.d. data. For RL, two challenges arise: (1) in many applications, the noise associated with a given variable is correlated across time, and hence no longer i.i.d.; (2) data are generated under a behavioral policy, potentially introducing distributional shifts and selection bias. Despite recent advances in CITs for stochastic processes \citep{liu2025conditional}, Markovian tests, and causal RL \citep{bareinboim2025intro_causal_rl}, open questions surround the adaptation of CITs to policy-intervened settings and, more generally, designing discovery algorithms that are robust in RL contexts. In this paper, 
we assume access to a ground truth longitudinal causal graph and explore (1) how it can be used to build a Markovian state and (2) whether that representation can be effectively leveraged by modern deep RL algorithms. 

\textbf{Causal Bandits and Causal Representation Learning}\;\; 
Causal bandits \citep{lattimore2016causal, bareinboim2015bandits, zhang2017transfer} focus on regret minimization in bandits, where actions causally affect the rewards via a (often known) causal graph \citep{sen2017identifying,lee2018structural,lee2019structural,lu2020regret,nair2021budgeted, lu2021causal}, and the environment resets after each round. 
Causal representation learning \citep{scholkopf2021toward, zhang2024causal, ahuja2023interventional, von2023nonparametric,varici2024general} aims to recover latent causal variables and their structural relationships from high-dimensional variables. Both address distinct settings orthogonal to the objective of this paper.
\section{Extended Preliminaries}
\label{appendix:prelims}
\textbf{Infinite Horizon Discounted Reward $W^{\text{th}}$ Order MDP\;\;} The only difference compared with episodic finite horizon $W^{\text{th}}$ order MDP lies in the objective. In infinite horizon discounted reward setting, we have an discount factor $\gamma$ and the objective is maximizing
\begin{equation*}
    \mathbb{E}^{\bm\pi}\left[\sum_{t=0}^\infty \gamma^tr(Z_h,A_h)\mid Z_t=z_t, A_t=a\right]
\end{equation*}
\section{Validity of I.I.D Noise}\label{sec:valid-iid-noise}
In this section, we establish that when the space of $X_t^i$ is countable, then if the SCM is describing an MDP, there exists a function $f^i_t$ such that $x_t^i=f_t^i(\mathbf{pa}_{X_t^i},u_t^i)$ where $u_t^i$ are independent uniform $[0,1]$. Without loss of generality, we assume there exists only one node per time step and the underlying MDP is 0-th order MDP, i.e. $\mathbb{P}(x_t|x_{t-1},\dots,x_0,a_{t-1})=\mathbb{P}(x_t|x_{t-1},a_{t-1})$. As stated in \cite{norris1998markov}, for any markov chain with countable state space, there exists an representation such that $x_t=f(x_{t-1},a_{t-1},u_t)$ where $U_t$ are i.i.d uniform random variable. Therefore, suppose the data forms an MDP, there must exist an causal function such that the noise is independent for countable state space.
\section{When is the Causal DAG  Helpful?}\label{sec:dag-helpful-causal-markov}
As discussed in Section~\ref{subsec:prelim} and shown in Theorem~\ref{thm:criterion_valid}, given DAG that satisfies Definition~\ref{def:markov_condition_dag} we can construct state such that it defines an MDP. However, when causal Markovian condition does not hold because of dependent noise, we can not construct state directly from DAG alone. For instance, suppose the DAG is just a single chain having $X_0\to X_1\to X_2\to X_3$ but the same noise $U$ controls all four nodes, we can not get that information from DAG and thus do not know whether just taking current observation state satisfies Markov property. We think it is an important future work to study how to construct valid state given DAG and other further information when causal Markovian condition fails. It is also important to study exact conditions on exogenous noise so that an SCM can admit an MDP. 
\section{Proof of Proposition~\ref{th:causal-markov-condition}}\label{sec:proof-causal-markov}
Before we prove Theorem~\ref{th:causal-markov-condition}, we introduce the notion of topological ordering which resorts the $\{\mathbf{X}_t\}_{t=0}^T$ so that the parents of $X^{(i)}$ can only be in $X^{(j)},j<i$. Formally,
\begin{definition}[Topological Ordering]
    Given a DAG $\mathcal{G}=(\mathbf{X},E)$, we say a mapping $g:\bm [T]\times[m]\to\{0,1,\dots,(T+1)(m+1)\}$ is a linear topological sort iff $\forall X_t^i,$ whenever $X_h^j\in\bm{pa}_{X_t^i}$, $g(h,i)\geq g(t,j)$.
\end{definition}
For any $X_t^i$, assume $\left|\textbf{nd}_{X_t^i}\right|=n$. Then we can topologically sort these nodes into $X^{(0)},X^{(1)},\dots,X^{(n)}$. Since $X_t^i$ does not have descendants among this set, without loss of generality, assume $X^{(n)}=X_t^i$. Then we have
\begin{align*}
    &\mathbb{P}\left(x^{(n)}\mid x^{(0)},x^{(1)},x^{(2)},\dots,x^{(n-1)}\right)\\&=\mathbb{P}\left(f^{(n)}\left(\mathbf{pa}_{X^{(n)}},U^{(n)}\right)\mid \mathbf{pa}_{X^{(n)}},U^{(0)},f^{(1)}\left(U^{(0)},U^{(1)}\right),\dots,f^{(n-1)}\left(U^{(0)},\dots, U^{(n-1)}\right)\right)\\
    &=\mathbb{P}(x^{(n)}\mid\mathbf{pa}_{X^{(n)}}),
\end{align*}
where the second equality holds because the variables are sorted by topological ordering and the last equality holds because of Assumption~\ref{assump:conditional-independence-noise}.
\section{Technical Details in Section~\ref{sec:dag-mdp}}
\label{sec:proofs}
\subsection{Proof of Theorem~\ref{thm:criterion_valid}}
We will first prove the sufficiency.
First, we prove that any state space satisfying the criterion satisfies the Markov property. It suffices to show that $\mathbb{P}(C_{t+1}^*=c_{t+1}, L_{t+1}=l_{t+1}\mid A_t=a,S_t=s_t,S_{t-1}=s_{t-1},\dots,S_0=s_0)=\mathbb{P}(C_{t+1}^*=c_{t+1}\mid A_t=a, S_t=s_t)$ where $\mathcal{L}_{t+1}=S_{t+1}\cap S_t$. This holds since $S_{t+1}\setminus S_t\subset \mathcal{C}_{t+1}^*$. Since $\mathcal{L}_{t+1}\subset S_t$, we have
\begin{align*}
    &\mathbb{P}(C_{t+1}^*=c_{t+1}, L_{t+1}=l_{t+1}\mid A_t=a,S_t=s_t,S_{t-1}=s_{t-1},\dots,S_0=s_0)\\&=\mathbb{I}(l_{t+1}=g(s_t))\mathbb{P}(C_{t+1}^*=c_{t+1}\mid A_t=a,S_t=s_t,S_{t-1}=s_{t-1},\dots,S_0=s_0),
\end{align*}
where $g$ is the projection from $S_t$ to $\mathcal{L_t}$. By construction of $\mathcal{C}_{t+1}^*$ and Criterion~\eqref{eq:state-parent-include}, for any $X\in\mathcal{C}_{t+1}^*$, $X$'s parent can only be in $\mathcal{C}_{t+1}^*\cup S_t$. Let $X_{t+1}^{(0)},\dots,X_{t+1}^{(i)},\dots$ be the variables in $\mathcal{C}_{t+1}^*$ that is ordered according to topological ordering. Then we have the parents of $X_{t+1}^{i}$ can only be $X_{t+1}^{(j)},j>i$ and $S_t$. Since by Criterion~\eqref{eq:state-inclusion-time}, if $S_{t+1}\cap {S}_{k}$ for $k<t$ is non empty, then those value are also contained in ${S}_t$, without losing generality, assume ${S}_{t+1}\cap{S}_k=\emptyset$ for $k<t.$ Then we have
\begin{align*}
    &\mathbb{P}(C_{t+1}^*=c_{t+1}\mid A_t=a,S_t=s_t,S_{t-1}=s_{t-1},\dots,S_0=s_0)\\
    &=\prod_{i=1}^{|\mathcal{C}_{t+1}^*|}\mathbb{P}\left(x_{t+1}^{(i)}\mid x_{t+1}^{(i-1)},\dots, x_{t+1}^{(1)},S_t=s_t,A_t=a,\dots,S_0=s_0\right)\\
&=\prod_{i=1}^{|\mathcal{C}_{t+1}^*|}\mathbb{P}\left(x_{t+1}^{(i)}\mid x_{t+1}^{(i-1)},\dots, x_{t+1}^{(1)},S_t=s_t,A_t=a\right),
\end{align*}
where the last equality holds because Criterion~\eqref{eq:state-decendent-non} ensures no path will point to $S_{t+1}$ without passing $S_t$.

Then we prove that such state space also preserves the optimal value. We prove by induction. The induction hypothesis is 
\begin{equation}\label{eq:indcution-hypothesis}
    \forall\, t, s_t,z_t,a, \;\; Q_t^*(s_t,a)=Q_t^*(z_t,a).
\end{equation}
When $t=T$, since by Criterion~\eqref{eq:reward-parent-include}, $\mathbf{pa}_{R_{T}}\subset \mathcal{S}_T$, we have $Q_T^*(s_T,a)=R(\mathbf{pa}_{R_{T}},a)=Q_T^*(z_t,a)$. Suppose this holds for $t=h+1$, when $t=h$, we have
\begin{align*}
    Q_h^*(s_h,a)&=R(s_h,a)+\mathbb{E}\left[\max_aQ^*_{h+1}(s_{h+1},a)\mid S_h=s_h,A_h=a\right]\\
    &=R(\mathbf{pa}_{R_h},a)+\mathbb{E}\left[\max_aQ^*_{h+1}(s_{h+1},a)\mid S_h=s_h,A_h=a\right].
\end{align*}
Similarly, we have
\begin{align*}
    Q_h^*(z_h,a)&=R(z_h,a)+\mathbb{E}\left[\max_aQ^*_{h+1}(z_{h+1},a)\mid Z_h=z_h,A_h=a\right]\\
    &=R(\mathbf{pa}_{R_h},a)+\mathbb{E}\left[\max_aQ^*_{h+1}(s_{h+1},a)\mid Z_h=z_h,A_h=a\right]\\
    &=R(\mathbf{pa}_{R_h},a)+\mathbb{E}\left[\max_aQ^*_{h+1}(s_{h+1},a)\mid S_h=s_h,A_h=a\right].
\end{align*}
where the second equality holds by Eq.~\eqref{eq:indcution-hypothesis} and the last equality holds because of Criterion~\eqref{eq:state-inclusion-time} and \ref{eq:state-parent-include}. Therefore, we have $Q_h^*(s_h,a)=Q_h^*(z_h,a)$. 

Next, we prove the necessity. To prove Criterion~\ref{eq:reward-parent-include},\ref{eq:state-parent-include} are necessary, we introduce Lemma~\ref{lem:parent_or_self}. The proof will be deferred to the end of this section.
\begin{lemma}[Necessity of parent inclusion]
\label{lem:parent_or_self}
Fix \(t \in \{0,\dots,T-1\}\). Suppose that the state variables 
\(\{\mathcal{S}_\tau\}_{\tau=t+1}^T\) have been constructed such that the following holds:
there exists a transition probability model consistent with the times series DAG under which, 
for any \(\tau \in \{t+1,\dots,T\}\), removing any variable from \(\mathcal{S}_\tau\) strictly reduces 
the optimal value.

Then, if there exists a variable \(X \in \mathcal{S}_{t+1}\cup\{R_t\}\) such that
\[
X \notin \mathcal{S}_t
\quad \text{and} \quad
\mathbf{pa}(C_{t+1}^*(X)) \cap \{ \text{variables at time } \le t \} \nsubseteq \mathcal{S}_t,
\]
then there exists a transition probability model consistent with the DAG under which
the optimal value at time \(t\) under \(\mathcal{S}_t\) is strictly smaller than that under the full construction.
\end{lemma}
Lemma~\ref{lem:parent_or_self} immediately shows that for any valid state, it must satisfy the first two criteria. 

Suppose Criterion~\ref{eq:state-inclusion-time} fails, there exists $X_h^i$ and $t_1<t_2<t_3$ such that $X_h^i\in S_{t_1},S_{t_3}, X_h^i\notin S_{t_2}$. Then we have $S_{t_3}$ is dependent of $S_{t_1}$ given $S_{t_2}$ since even conditioned on $S_{t_2}$, $X_h^i$ in $S_{t_1}$ determines the value of $X_h^i$ in $S_{t_3}$. This breaks the Markovian property.

Since Criterion~\ref{eq:state-inclusion-time} must hold, it implies
$
N_{t+1}\cap\left(\bigcup_{h=0}^{t}S_h\right)=\varnothing.
$
Indeed, if a variable in $N_{t+1}$ had appeared in some $S_h$ with
$h<t$, then its appearances in $S_h$ and $S_{t+1}$ would imply its
inclusion in $S_t$ by persistence. This contradicts the definition
of $N_{t+1}$.

Suppose Criterion~\ref{eq:state-decendent-non} fails, there exists $t$ and $s\le t$, maximal clique $K$ of $M_t$
satisfying $K\cap N_{t+1}\neq\varnothing$, $X_s^k$ satisfying $X_s^k\in A_{t,s}$ and a variable $X_r^j$ such that
\[
X_r^j
\in
\bigl(\{X_s^k\}\cup\mathbf{pa}(X_s^k)\bigr)
\cap
\left[
\left(\bigcup_{h=0}^{t}S_h\right)\setminus S_t
\right],
\]
and
\[
\bigl(\{X_s^k\}\cup\mathbf{pa}(X_s^k)\bigr)\cap K
\neq\varnothing.
\]
Choose
$
X_h^i\in K\cap N_{t+1}
$
and
$
X_\ell^m
\in
\bigl(\{X_s^k\}\cup\mathbf{pa}(X_s^k)\bigr)\cap K.
$
Thus $X_r^j$ is a previously included variable that is absent
from $S_t$, and it shares a family with a variable in $K$.

Consider the subgraph of $\mathcal G$ induced by
$
N_{t+1}\cup\left(\bigcup_{h=0}^{t}S_h\right)
$
and all ancestors of these variables, and form its moral graph.
Recall that moralization connects every pair of parents sharing
a child and then removes all arrowheads.

Every variable in $A_{t,s}$ belongs to this ancestral subgraph.
To see this, the variables used to define $A_{t,s}$ are indexed
by $s\le t$ and belong to states up to time $t+1$. Each such
variable therefore either belongs to
$\bigcup_{h=0}^{t}S_h$ or belongs to $N_{t+1}$.
The same-time ancestors added by $C_s^*$, and all their parents,
also belong to the ancestral subgraph.

Consequently, every edge of $M_t$ is also present in this
ancestral moral graph. Furthermore,
\[
\{X_s^k\}\cup\mathbf{pa}(X_s^k)
\]
forms a clique there: the child is adjacent to each parent,
and every pair of parents is joined by moralization.
In particular, the ancestral moral graph contains an edge
between $X_\ell^m$ and $X_r^j$.

Because $K$ is a clique, it is connected. Hence there is a path
inside $K$ from $X_h^i$ to $X_\ell^m$; if these variables
coincide, this path has length zero. Appending the edge from
$X_\ell^m$ to $X_r^j$ gives a path from $X_h^i$ to $X_r^j$
in the ancestral moral graph.

Every vertex of $K$ lies outside $\bigcup_{h=0}^{t}S_h$,
by the construction of $M_t$. Therefore, every vertex of
the path inside $K$ lies outside $S_t$. Its final vertex
$X_r^j$ also lies outside $S_t$ by its choice.
Thus this path avoids $S_t$ entirely.

It follows that $S_t$ does not separate $N_{t+1}$ from
\[
\left(\bigcup_{h=0}^{t}S_h\right)\setminus S_t
\]
in their ancestral moral graph. By the ancestral moral graph
criterion for d-separation,
\[
N_{t+1}
\not\!\perp_d
\left[
\left(\bigcup_{h=0}^{t}S_h\right)\setminus S_t
\right]
\mid S_t.
\]

By completeness of d-separation, there exists a probability
model compatible with $\mathcal G$ under which
\[
N_{t+1}
\not\!\perp\!\!\!\perp
\left[
\left(\bigcup_{h=0}^{t}S_h\right)\setminus S_t
\right]
\mid S_t.
\]
However, $N_{t+1}\subseteq S_{t+1}$, whereas the Markov property
requires
\[
S_{t+1}
\perp\!\!\!\perp
(S_0,\ldots,S_{t-1})
\mid S_t.
\]
The latter would imply the conditional independence just shown
to fail. This contradicts Markovianity for every compatible model. Therefore, the required inclusion holds.
\begin{algorithm}[!h]
\caption{Backward State-Space Construction from a Time Series DAG}
\label{alg:backward_state_construction}
\begin{algorithmic}[1]
\REQUIRE Time horizon \(T\); times series DAG \(\mathcal{G}\)\;
\STATE Initialize ${S}_{T+1} \gets \emptyset$, 
$
\mathcal{C}_{T+1}\gets\emptyset$
\FOR{\(t = T, T-1, \dots, 0\)}
    \STATE Construct state $S_t$ and auxiliary variable set $\mathcal{C}_t$:
   \begin{align*}
    S_t
    &\gets
    \left
    \{X_s^i: X_s^i\in S_{t+1},
    s\leq t\right\} \cup 
    \{
    X_s^i:
    X_s^i \in \mathbf{pa}_{X_{t+1}^j},
    X_{t+1}^j \in S_{t+1}
    \cup\mathcal{C}_{t+1}
    ,
    s \le t
    \}
    \cup \mathbf{pa}_{R_t}
   \end{align*}

    \STATE Initialize $
    \mathcal{C}_t
    \gets
    \left\{
    X_t^i \;:\;
    X_t^i \in \mathbf{pa}_{X_t^j},\;
    X_t^j \in S_t,
    \; X_t^i\notin S_t
    \right\}.$\;
    \WHILE{$\left\{X_t^i \;: \; X_t^i\in\mathbf{pa}_{X_t^j},\;X_t^j\in\mathcal{C}_t
    ,\;X_t^i\notin \mathcal{C}_t
    \right\}\neq \emptyset$}
    \STATE $\mathcal{C}_t\gets\mathcal{C}_t\cup \left\{X_t^i\;:\; X_t^i\in\mathbf{pa}_{X_t^j},\;X_t^j\in\mathcal{C}_t,\;X_t^i\notin \mathcal{C}_t\right\}$
\;
\ENDWHILE

\ENDFOR

\STATE \textbf{Output:} \(\{\mathcal{S}_t\}_{t=0}^T\)
\end{algorithmic}
\end{algorithm}



\subsection{Proof of Theorem~\ref{thm:minimality}}

We first show that this construction gives a valid state space by verifying the four criteria of Theorem~\ref{thm:criterion_valid}. Criteria~\eqref{eq:reward-parent-include} and \ref{eq:state-parent-include} is satisfied by Line 3 of Algorithm~\ref{alg:backward_state_construction} since $\mathcal{C}_{t+1}\cup\mathcal{S}_{t+1}$ is the same as $\mathcal{C}_{t+1}^*$. For Criterion~\eqref{eq:state-decendent-non}, again by Line 3, we have for all $k\leq t_2$ if $X_k^i\in S_{t_1}$, $X_k^i\in S_{t_2}$. Therefore, for every $X_k^i\in S_{t_1}\setminus S_{t_2}$, time index $k$ must be larger than $t_2$. Since the variables in $S_{t_2}$ can only have ancestors from time smaller or equal to $t_2$, Criterion~\eqref{eq:state-decendent-non} is satisfied. Criterion~\eqref{eq:state-inclusion-time} holds because of Line 3 similarly.

We will use induction to show that we can't move any variable from the state constructed by Algorithm~\ref{alg:backward_state_construction}. At $t=T$, by Lemma~\ref{lem:parent_or_self}, we must have the parents of reward to preserve optimality, which is exactly the same as the construction of Algorithm~\ref{alg:backward_state_construction}. Suppose for $t\geq h+1$, we can't remove any variable from the constructed state. When $t=h$, by Lemma~\ref{lem:parent_or_self}, for any $X\in\mathcal{S}_{h+1}$, either itself or $\mathbf{pa}(C_{h+1}^*(X))$ needs to be in $\mathcal{S}_{h}$. If $X$ is indexed by $k\leq h$, $X$ can be in the state at time $h$. Since this will only increase the dimension of state space by 1 which is smaller than the cardinality of $\mathbf{pa}(C_{h+1}^*(X))$, we should put $X$ to $\mathcal{S}_t$. If $X$ indexed by $h+1$, then $\mathbf{pa}(C_{h+1}^*(X))$ must be in $\mathcal{S}_t$. 




Since this exactly matches the output of Algorithm~\ref{alg:backward_state_construction}, every variable retained by the construction is indispensable: deleting any one of them causes failure of at least one of the two desired properties for some transition instance.

Therefore, no strictly smaller state sequence can simultaneously preserve Markovianity and preserve the optimal policy for all transition instances. This proves the minimality of \(\{\mathcal{S}_t\}_{t=0}^T\).
\subsection{Proof of Lemma~\ref{lem:parent_or_self}}
It suffices to consider the case where exactly one necessary variable $X$ is omitted from the state at time $t$. Suppose $X$ is either a causal parent of some variable in $S_{t+1}$ with time index at most $t$, or a variable in $S_{t+1}$ itself, but $X \notin S_t$. Fix some realization $S_t=s$, and let $X\in\{0,1\}$ with
\begin{equation}
\mathbb{P}(X=1)=p,\qquad \mathbb{P}(X=0)=1-p,
\end{equation}
for some $p\in(0,1)$. Consider two actions $a_0$ and $a_1$, and define a transition model consistent with the DAG as follows: the next state contains a binary variable $Y_{t+1}\in S_{t+1}$, and
\begin{equation}
\mathbb{P}(Y_{t+1}=1\mid S_t=s,X=x,A_t=a)
=
\mathbb{I}(a=a_x),
\end{equation}
where $a_x=a_0$ if $x=0$ and $a_x=a_1$ if $x=1$. Let the continuation value be
\begin{equation}
V_{t+1}(Y_{t+1})=\mathbb{I}(Y_{t+1}=1).
\end{equation}
Then if $X$ were observed as part of the state at time $t$, the controller could choose $a_x$ after observing $x$, and hence attain value
\begin{equation}
\max_a \sum_{y} \mathbb{P}(Y_{t+1}=y\mid S_t=s,X=x,A_t=a)V_{t+1}(y)=1
\end{equation}
for both $x=0$ and $x=1$. In contrast, if $X$ is not included in the state, then the policy must choose the same action for both values of $X$, so the value is obtained by averaging over the conditional law of the unobserved variable:
\begin{equation}
Q_t(s,a)
=
\sum_{x\in\{0,1\}} \mathbb{P}(X=x\mid S_t=s)\sum_y \mathbb{P}(Y_{t+1}=y\mid S_t=s,X=x,A_t=a)V_{t+1}(y).
\end{equation}
By construction,
\begin{equation}
Q_t(s,a_0)=1-p,\qquad Q_t(s,a_1)=p,
\end{equation}
and therefore
\begin{equation}
\max_a Q_t(s,a)=\max\{p,1-p\}<1
\end{equation}
whenever $p\in(0,1)$. Thus, the optimal value under the reduced state representation is strictly smaller than under the full state representation. Moreover, any policy based only on the reduced state $S_t=s$ must choose the same action regardless of the realized value of $X$, whereas the optimal full-state policy chooses $a_0$ when $X=0$ and $a_1$ when $X=1$. Hence the policy induced by the reduced state representation is not optimal under the full state representation. This proves that omitting even a single necessary variable can strictly reduce the optimal value.
    

\begin{algorithm}[!t]
\caption{Multi-Order State Exposure (MOSE)}
\label{alg:mose}
\begin{algorithmic}[1]
\REQUIRE \textbf{Update-Value-Function};
order $W$; batch size $B$; head dimension $d$

\STATE Initialize separate parameters $\theta_h$ for each
$h=1,\ldots,H$, including the attention module and Q-network
\STATE Initialize thresholds $\tau_h=0$ and replay buffers
$\mathcal{D}_h=\emptyset$

\FOR{$k=1,2,\ldots,K$}
    \STATE Reset the environment
    \FOR{$h=1,2,\ldots,H$}
        \STATE Observe $\mathbf{X}_h^k$ and construct
        $Z_h^k=
        (\mathbf{X}_h^k,\ldots,\mathbf{X}_{h-W}^k)$\;
        \STATE Construct $\hat{O}_h^k$ using the step-$h$
        attention and selection operations defined below\;
        \STATE Take action
        $a_h^k=\arg\max_a
        Q_h^{\theta_h}(\hat{O}_h^k,a)$
        \STATE Observe $r_h^k$ and $O_{h+1}^k$\;
        \STATE Store
        $(Z_h^k,a_h^k,r_h^k,O_{h+1}^k)$
        in $\mathcal{D}_h$
    \ENDFOR

    \FOR{$h=H,H-1,\ldots,1$}
        \STATE Sample $B$ transitions
        $\{(Z_h^{k_b},a_h^{k_b},r_h^{k_b},
        O_{h+1}^{k_b})\}_{b=1}^{B}$
        from $\mathcal{D}_h$

        \FOR{$b=1,\ldots,B$}
            \STATE Let $\mathbf{A}_h$ denote the availability
            mask and $\mathcal{I}_h$ the available coordinates

            \STATE
            $E_{h;t,i}^{b}
            =\tanh\!\left(
            W_h^i X_t^{k_b,i}+b_{h;h-t,i}
            \right)$
            \STATE  $\mathbf{E}_h^b:=(E_{h;t,i}^{b})$\;
           
                \STATE Compute keys and values:
                $\mathbf{K}_h^{b}
                =\mathbf{E}_h^b W_{K,h}$,
                $\mathbf{V}_h^{b,j}
                =\mathbf{E}_h^b W_{V,h}$

                \STATE Compute attention weights:
                $\boldsymbol{\alpha}_h^{b}
                =\operatorname{softmax}_{\mathcal{I}_h}
                \!\left(
                \frac{\mathbf{q}_{0,h}
                (\mathbf{K}_h^{b})^\top}{\sqrt{d}}
                \right)$

                \STATE Compute context:
                $\mathbf{c}_h^{b}
                =\boldsymbol{\alpha}_h^{b}
                \mathbf{V}_h^{b}$

            \STATE Compute coordinate scores:
            $\mathbf{z}_h^b
            =\dfrac{1}{\sqrt{d}}
            \mathbf{c}_h^{b}
            (\mathbf{K}_h^{b})^\top$

            \STATE Select coordinates:
            $\mathbf{m}_h^b
            =\mathbf{A}_h\odot
            \mathbf{1}\!\left\{
            \sigma(\mathbf{z}_h^b)\geq\tau_h
            \right\}$

            \STATE Construct the exposed state:
            $\hat{O}_h^{k_b}
            =[\,Z_h^{k_b}\odot\mathbf{m}_h^b;
            \mathbf{m}_h^b\,]$
        \ENDFOR

        \STATE $\displaystyle
        \theta_h\gets
        \textbf{Update-Value-Function}\!\left(
        \theta_h,
        \left\{
        (\hat{O}_h^{k_b},a_h^{k_b},
        r_h^{k_b},O_{h+1}^{k_b})
        \right\}_{b=1}^{B}
        \right)$
    \ENDFOR
\ENDFOR
\end{algorithmic}
\end{algorithm}
\begin{algorithm}[!t]
\caption{Causal MOSE}
\label{alg:causal-mose}
\begin{algorithmic}[1]
\REQUIRE \textbf{Update-Value-Function}; $W,B,d$;
minimal Markovian state masks $\{\mathbf{m}_h^\star\}_{h=1}^{H}$

\STATE Initialize as in MOSE, with $\theta_h=(\phi_h,\psi_h)$:
$\phi_h$ contains attention and selection parameters;
$\psi_h$ contains Q-network weights

\FOR{$k=1,\ldots,K$}
    \STATE Collect transitions as in MOSE
    \FOR{$h=H,\ldots,1$}
        \STATE Sample $B$ transitions $\mathcal{B}_h$
        from $\mathcal{D}_h$
        \STATE $(\theta_h,\mathbf{y}_h)\gets
        \textbf{MOSE}(h,\theta_h,\mathcal{B}_h)$
        \COMMENT{One minibatch update; retain detached TD targets}
        \STATE $\psi_h\gets
        \textbf{DAG-State-Update}
        (\psi_h,\mathcal{B}_h,\mathbf{m}_h^\star,\mathbf{y}_h)$
    \ENDFOR
\ENDFOR
\end{algorithmic}
\end{algorithm}

\begin{algorithm}[!t]
\caption{DAG-State-Update}
\label{alg:dag-state-update}
\begin{algorithmic}[1]
\REQUIRE Q-network weights $\psi_h$; batch $\mathcal{B}_h$;
minimal-state mask $\mathbf{m}_h^\star$;
detached TD targets $\mathbf{y}_h$

\FOR{$b=1,\ldots,B$}
    \STATE $O_h^{\star,k_b}
    =[\,Z_h^{k_b}\odot\mathbf{m}_h^\star;
       \mathbf{m}_h^\star\,]$
\ENDFOR

\STATE Bypass attention and selection; compute
\[
\mathcal{L}_h^{\mathrm{DAG}}
=
\frac{1}{B}\sum_{b=1}^{B}
\left(
Q_h^{\psi_h}(O_h^{\star,k_b},a_h^{k_b})-y_h^b
\right)^2
\]
\STATE Take one optimizer step on $\psi_h$ using
$\nabla_{\psi_h}\mathcal{L}_h^{\mathrm{DAG}}$;
keep $\phi_h$ fixed
\STATE \textbf{return} $\psi_h$
\end{algorithmic}
\end{algorithm}
\section{Experimental Details}
\label{appendix:experimental_details}

\textbf{Compute Resources \;}  Experiments were run on an AWS EC2 c6a.4xlarge instance with 16 vCPUs.


\subsection{Synthetic DGP Details}\label{app:DGP_details}
In the linear additive case, each variable takes the form 
\begin{equation}\label{eq:linear-anm}
    X_t^i=b^i+\omega^{i}_{t,a}a+\sum_{x_{h}^j\in\mathbf{pa}(X_t^i)}\omega_{t,x_{h,j},a}^ix_{h}^j+\epsilon_{t}^i,
\end{equation}
where $\omega$ is randomly generated, $b^i$ follows normal distribution with mean 0 and $\epsilon_t^i$ is uniform over $[-\sqrt{3},\sqrt{3}]$. And the reward follows the same DGP.

Additionally, we have implemented a non-linear ANM. 
In the nonlinear additive case, we standardize the parent values as
$z_h^j=(x_h^j-\mu_h^j)/s_h^j$ and $z_a=(a-\mu_a)/s_a$,
where $\mu$ and $s>0$ are randomly generated parameters whose
dependence on the target node $X_t^i$ is suppressed.
The action modifies each state-parent coefficient as
\begin{equation*}
\omega_{t,x_{h,j},a}^i
=\omega_{t,x_{h,j}}^i+\Delta_{t,x_{h,j},a}^i.
\end{equation*}
Define $\rho(z)=z/(1+|z|)$ and the weighted nonlinear functions
\begin{equation*}
\begin{aligned}
\varphi_1(z,w)&=w\tanh(z),&
\varphi_2(z,w)&=w\max\{0,\operatorname{sgn}(w)z\},\\
\varphi_3(z,w)&=w[\sin(z+\theta_t^i)
                    +\tfrac12\cos(\nu_t^i z)],&
\varphi_4(z,w)&=\rho(wz).
\end{aligned}
\end{equation*}
Then define
\begin{equation*}
f_{m,t,a}^i
=\varphi_m(z_a,\omega_{t,a}^i)
+\sum_{x_h^j\in\mathbf{pa}_{X_t^i}}
 \varphi_m(z_h^j,\omega_{t,x_{h,j},a}^i),
\qquad m=1,\ldots,4.
\end{equation*}
Pairwise interactions and a sigmoid gate are defined by
\begin{equation*}
f_{5,t,a}^i
=\sum_{\{x_h^j,x_u^k\}\subseteq\mathbf{pa}_X(X_t^i)}
 \eta_{t,h,u}^{i,j,k}\rho(z_h^jz_u^k),
\end{equation*}
\begin{equation*}
f_{6,t,a}^i
=\sigma_{20}\!\left(
 \gamma_{t,a}^i z_a+
 \sum_{x_h^j\in\mathbf{pa}_{X_t^i}}
 \gamma_{t,h,j}^i z_h^j\right),
\qquad
\sigma_{20}(z)=
\frac{1}{1+\exp\{-\operatorname{clip}_{[-20,20]}(z)\}}.
\end{equation*}
The pairwise sum counts each distinct unordered pair once.
Each node randomly selects a coefficient vector $c_t^i$ from its
eligible mixtures in the fixed library, giving
\begin{equation*}
f_{t,a}^i
=\sum_{m=1}^{5}c_{m,t}^i f_{m,t,a}^i
+c_{6,t}^i f_{6,t,a}^i f_{5,t,a}^i
+c_{7,t}^i\rho(f_{5,t,a}^i)
+c_{8,t}^i f_{6,t,a}^i f_{4,t,a}^i.
\end{equation*}
Each variable then takes the form
\begin{equation}\label{eq:nonlinear-anm}
X_t^i=
\operatorname{clip}_{[-10^6,10^6]}
\left[b^i+g_t^i f_{t,a}^i
+\alpha_t^i\sin(\nu_t^i t+\theta_t^i)\right]
+\epsilon_t^i.
\end{equation}
The weights $\omega$, action adjustments $\Delta$, interaction
coefficients $\eta$, gate coefficients $\gamma$, and parameters
$g,\alpha,\nu,\theta$ are randomly generated.
The intercept $b^i$ follows a normal distribution with mean zero,
and $\epsilon_t^i$ is independently drawn from
$\operatorname{Unif}(-\sqrt{3},\sqrt{3})$.
Reward nodes follow the same construction, with the clipped component
divided by $1000$ and an action-dependent offset added before the noise.
\subsection{Ablations on Attention}\label{sec:ablation-attention}
\begin{figure}
    \centering
    \includegraphics[width=\linewidth]{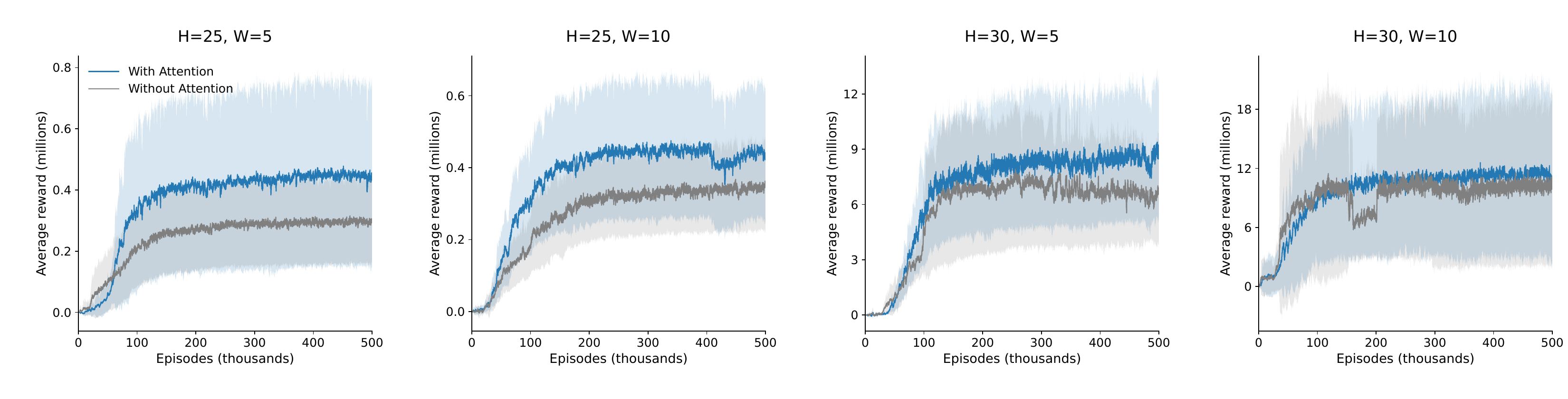}
    \caption{Performance Comparison Between Using Attention to Score and Directly Learn Score. Averaged over 18 instances. }
    \label{fig:ablation-attention}
\end{figure}

We compared MOSE with a version without attention. Specifically, instead of using attention to calculate $s$ and then compare with learnable threshold $\tau$, we directly set $s$ as learnable parameter for each dimension. As shown in Figure~\ref{fig:ablation-attention}, MOSE without attention is worse than MOSE with attention. This indicates that with the learnable query, it can more directly calculates the relationship between the estimation error and each dimension of history. 
\subsection{Per Horizon Loss Difference Curve}\label{sec:per-horizon-loss-difference}
\begin{figure}
    \centering
    \includegraphics[width=\linewidth]{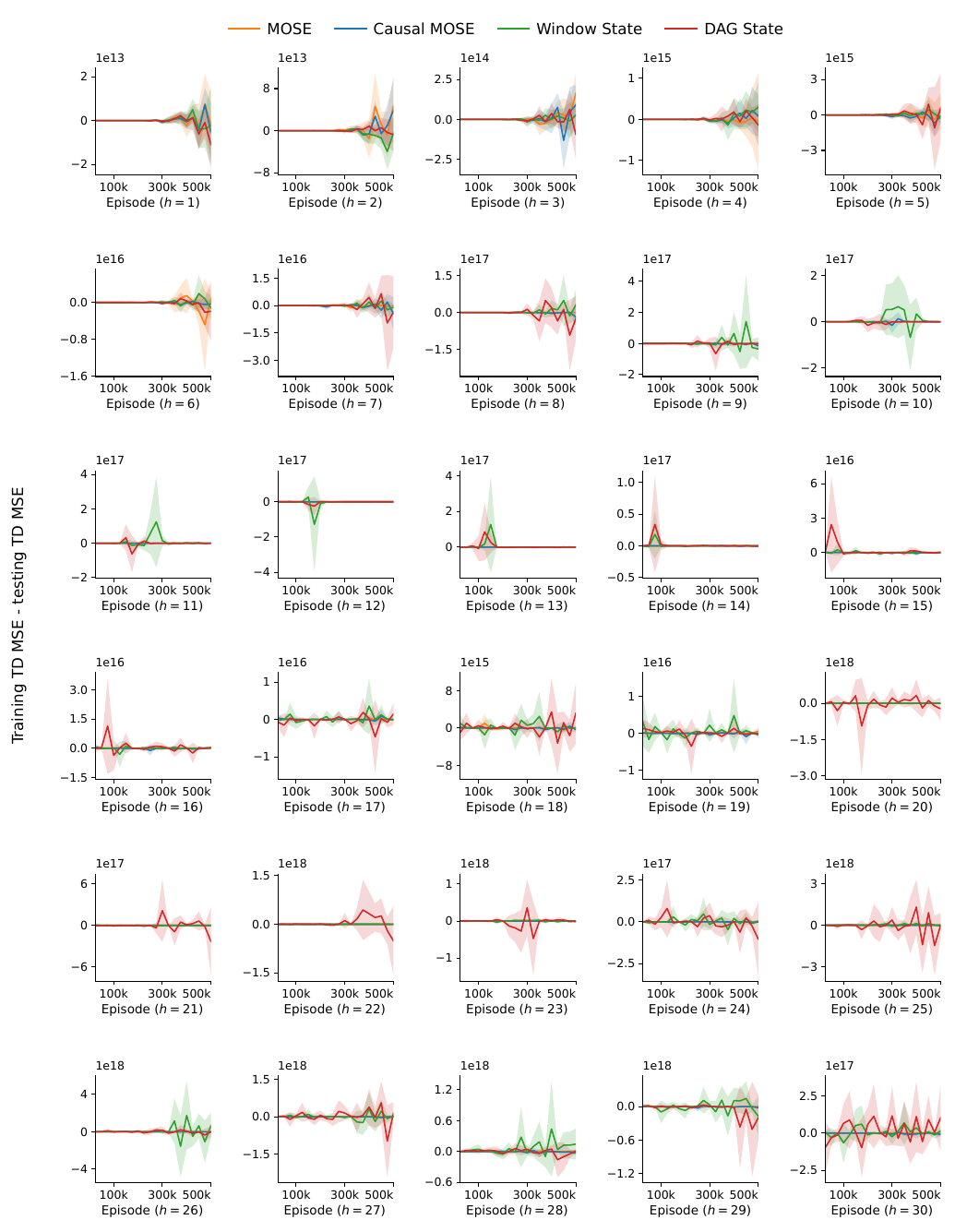}
    \caption{Difference of Training and Testing Loss for Horizon 30 and $W=5$}
    \label{fig:difference-30-5-10}
\end{figure}

\begin{figure}
    \centering
    \includegraphics[width=\linewidth]{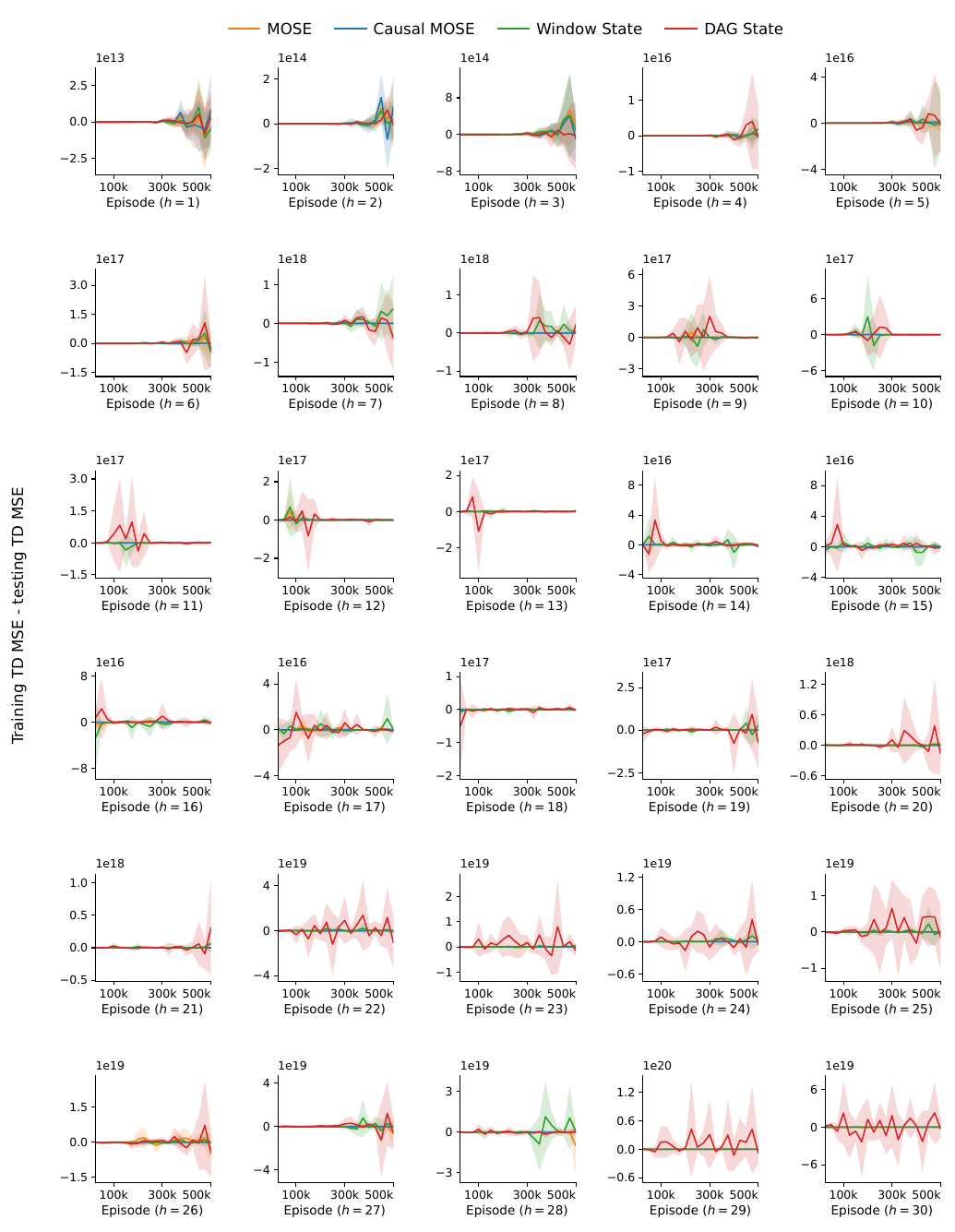}
    \caption{Difference of Training and Testing Loss for Horizon 30 and $W=10$}
    \label{fig:difference-30-10}
\end{figure}

As shown in Figures~\ref{fig:difference-30-5-10}-\ref{fig:difference-30-10}, many curves are initially flat and close to zero, indicating little discrepancy between training and testing losses on the early samples. Their subsequent departures from zero are consistent with our interpretation of overfitting to early samples: a model may generalize well within the early sample distribution but become overly adapted to it, limiting its ability to accommodate samples encountered later in training. For Figure~\ref{fig:difference-30-5-10}, this deterioration is particularly visible for DAG State at $h=20$--$25$, whereas MOSE and Causal MOSE remain comparatively close to zero. Window State also develops substantial fluctuations at $h=26$ and $h=28$, suggesting that a fixed history representation can exhibit similar difficulties. For Figure~\ref{fig:difference-30-10}, the contrast is especially pronounced at $h=22$--$25$ and $h=29$--$30$, where DAG State exhibits large fluctuations while Causal MOSE remains comparatively stable. MOSE also performs well across many of these steps, although it develops late-training deviations at $h=26$ and $h=28$. At the earliest horizon steps, the advantage is less consistent, as the MOSE variants also depart from zero later in training. These patterns suggest that learned state selection can help mitigate excessive adaptation to early samples, particularly at intermediate and late horizon steps.
\subsection{Nonlinear ANM (Eq.~\ref{eq:nonlinear-anm})}
\begin{figure}
    \centering
    \includegraphics[width=\linewidth]{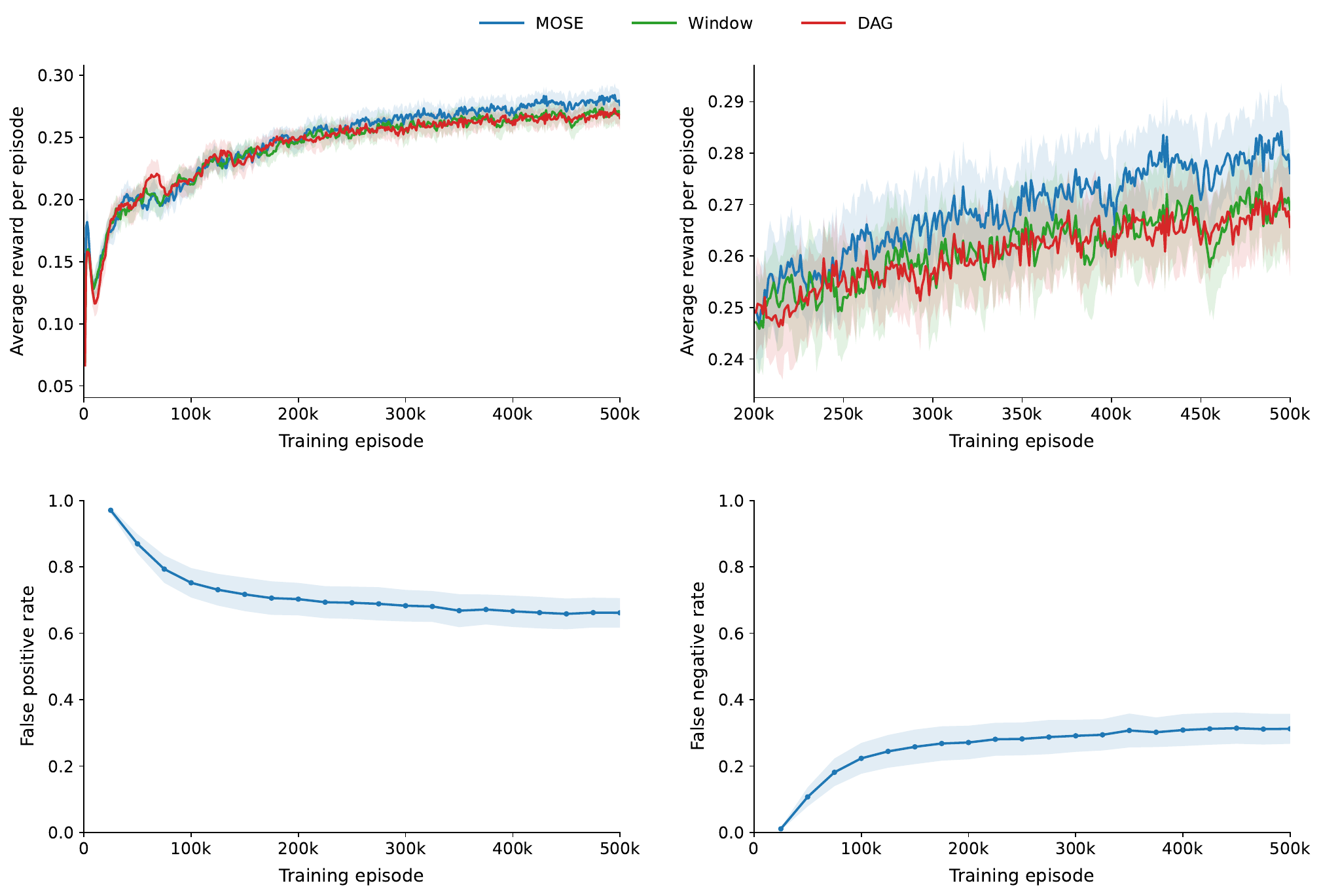}
    \caption{Avergaed over 18 instances of nonlinear ANM. Horizon is 30 and $W=10$. 20 nodes per time step. Top: Performance comparison; Bottom: False positive/negative rate of MOSE selected state.}
    \label{fig:nonlinear-anm}
\end{figure}
As shown in the top row of Figure~\ref{fig:nonlinear-anm}, the closely clustered reward curves and small absolute performance gaps suggest that obtaining a clear advantage in this complex-transition setting is challenging. Nevertheless, MOSE eventually achieves higher mean rewards than both the Window and DAG state benchmarks, as highlighted in the upper-right panel. This result suggests that MOSE can capture useful information for decision making even under complicated transition dynamics. The second row shows that its false-positive rate decreases during training, although its false-negative rate increases. Thus, the performance gains support the usefulness of MOSE’s learned representation, while the gate errors indicate that it does not exactly recover the DAG-defined state due to finite samples.
\subsection{What Is MOSE Learning?}\label{sec:mose-dag}
\begin{figure}
    \centering
    \includegraphics[width=\linewidth]{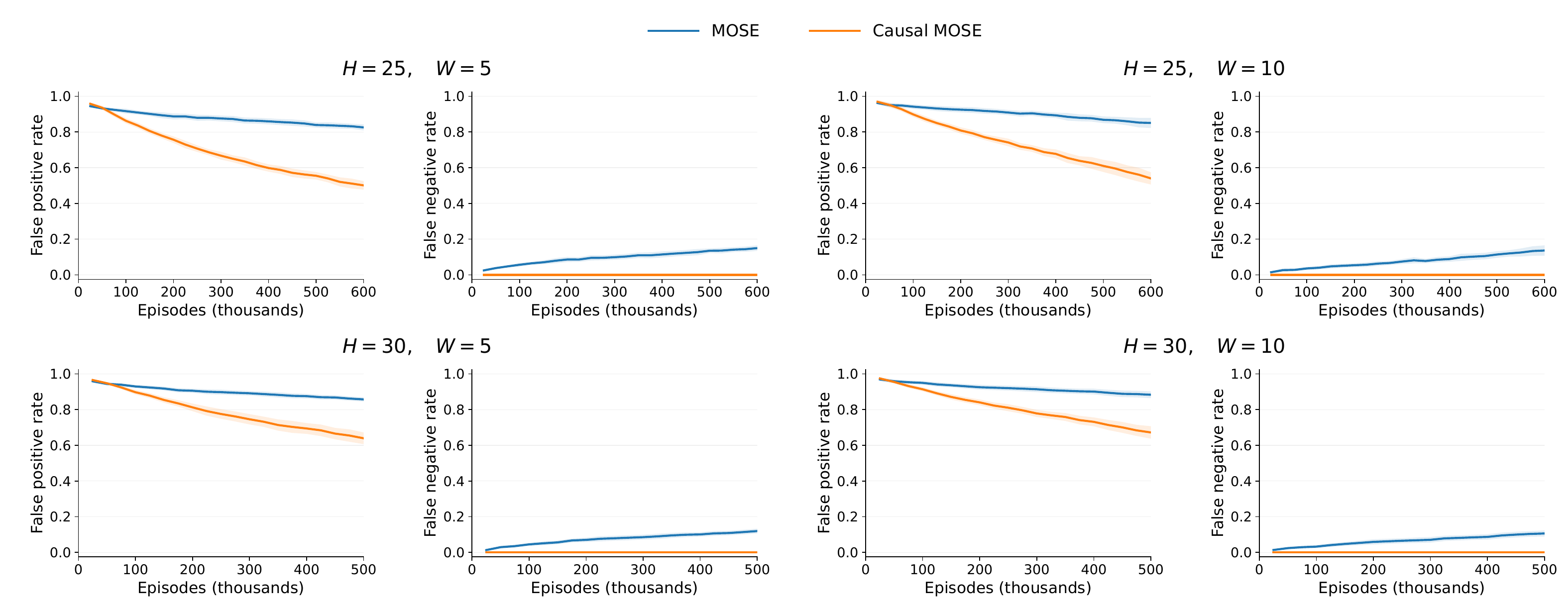}
    \caption{False positive/negative rate of MOSE selected state and Causal MOSE selected state.}
    \label{fig:false-positive-anm}
\end{figure}
As shown in the Figure~\ref{fig:false-positive-anm} and bottom row of Figure~\ref{fig:nonlinear-anm}, MOSE progressively eliminates some historical coordinates and retains a smaller subset, as indicated by its decreasing false-positive rate. However, its increasing false-negative rate raises the question of whether this selection recovers a minimal Markovian state. First, the minimal state constructed by Algorithm~\ref{alg:backward_state_construction} does not imply uniqueness without additional assumptions. For a simple counterexample, let $X_t$ be a nondegenerate binary Markov chain, observe both $X_t$ and $Y_t = 1-X_t$, and define the reward as $R_t = \mathbf{1}\{A_t = X_t\}$. Both $\{X_t\}$ and $\{Y_t\}$ are minimal Markovian states: either determines the transition and reward distributions, whereas the empty state cannot preserve the optimal action. Thus, omitting a coordinate from one particular minimal state does not necessarily imply losing essential information. 

Second, Causal MOSE consistently outperforms MOSE in the available state-selection comparisons and eliminates unnecessary coordinates more rapidly, supporting the usefulness of the DAG state constructed by our algorithm in guiding selection. In these runs, DAG coordinates are retained by construction and this results in faster reduction in false positives, demonstrating improved elimination of additional information by having minimal markovian state information. Together, these observations are consistent with finite-sample estimation and optimization errors causing MOSE to discard coordinates belonging to the algorithm's minimal Markovian state. They motivate the conjecture that, with sufficient data and suitable learning conditions, MOSE may recover the DAG state or an informationally equivalent minimal state asymptotically.
\subsection{Learning Curve of Atari Games}\label{sec:learning-curve-atari}
\begin{figure}
    \centering
    \includegraphics[width=\linewidth]{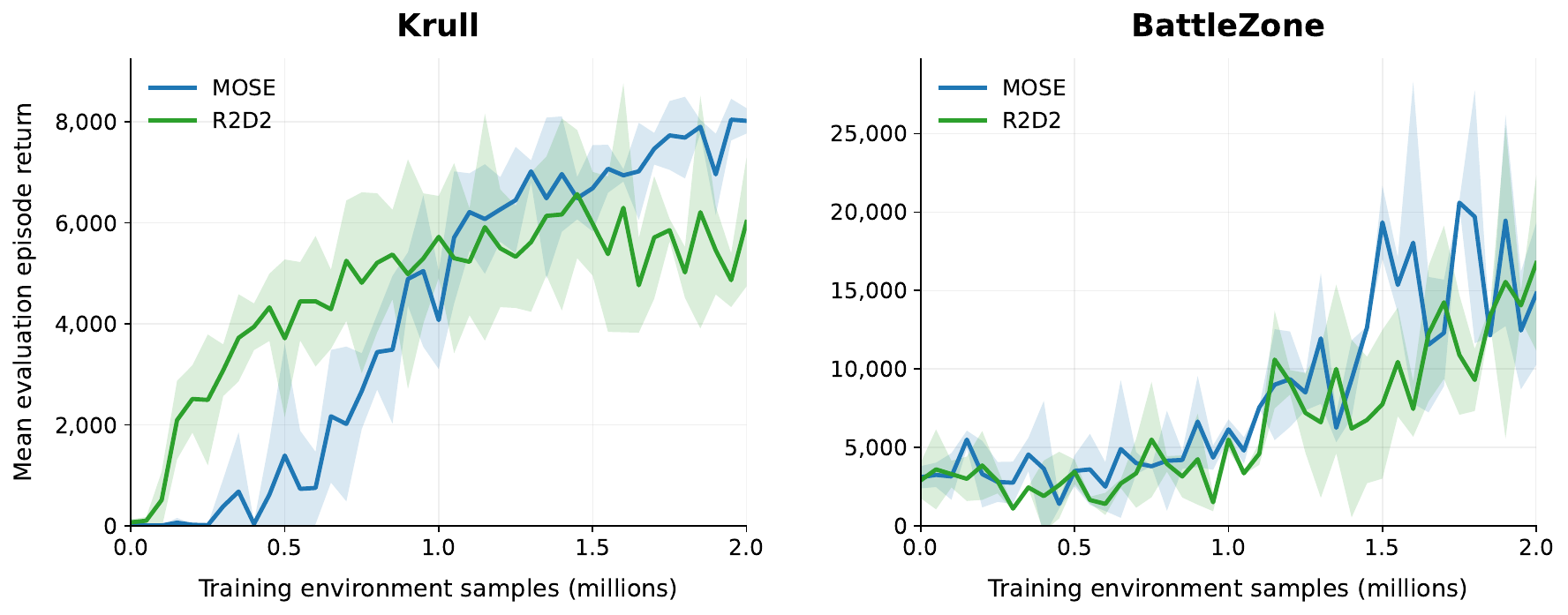}
    \caption{Performance Comparison on Krull and Battlezone}
    \label{fig:atari-games}
\end{figure}

As shown in Figure~\ref{fig:atari-games}, MOSE frequently outperforms R2D2 on BattleZone, particularly during later training. On Krull, MOSE requires a longer initial learning period but subsequently overtakes R2D2 and sustains higher returns toward the end of training. This delayed yet sustained improvement suggests that explicit selection may help MOSE eventually learn a more informative belief state.
\subsection{Choices of Hyperparameters}
We report the hyperparameters of Q-netwrok for synthetic and the attention related hyperparameters for Atari games respectively in Table~\ref{tab:hyperparams} and \ref{tab:hyperparams-gopher}. For the choices of hyperparameters of R2D2, we match the ones reported in \cite{kapturowski2018recurrent} except we only have one learner that collects sample and learns from it. We did not have multiple actors as in R2D2 due to compute constraint.
\begin{table}[t]
\centering
\caption{Hyperparameters of Network for Synthetic}
\label{tab:hyperparams}
\begin{tabular}{ll}
\hline
Hyperparameter & Value \\
\hline
Hidden layers & 2 \\
Hidden size & 128 \\
Learning rate & $10^{-3}$ \\
Adam $\epsilon$ & $10^{-8}$ \\
Batch size & 64 \\
Replay capacity & 50{,}000 \\
Target update interval & 50 \\
Replay frequency & 1 \\
Gradient clipping & 1000.0 \\
Learning starts & 32 \\
\hline
\end{tabular}
\end{table}
\begin{table}[t]
\label{tab:hyperparams-gopher}
\centering
\caption{MOSE attention selection-layer hyperparameters.}
\begin{tabular}{lc}
\hline
Hyperparameter & Value \\
\hline
History length & 4 \\
Attention dimension & 32 \\
\hline
\end{tabular}
\end{table}


\end{document}